\documentclass{article}

\usepackage{arxiv}

\usepackage[utf8]{inputenc} 
\usepackage[T1]{fontenc}    
\usepackage{hyperref}       
\usepackage{url}            
\usepackage{booktabs}       
\usepackage{amsfonts}       
\usepackage{nicefrac}       
\usepackage{microtype}      
\usepackage{lipsum}		
\usepackage{graphicx}
\usepackage{natbib}
\usepackage{doi}
\usepackage{xcolor}
\usepackage{hyperref}
\usepackage{amssymb}
\usepackage{amsmath}
\usepackage{amsthm}
\usepackage{subcaption}
\usepackage{enumitem}
\usepackage{tabularx}
\usepackage{array}
\usepackage{booktabs}
\usepackage{graphicx}

\newcommand{\vpara}[1]{\paragraph{#1 }}
\newcommand{\aria}[1]{TheoryTab}

\title{Computational Copyright: Towards A Royalty Model for Music Generative AI}

\date{} 					

\author{%
  Junwei Deng\textsuperscript{\textnormal{1}}\quad
  Xirui Jiang\textsuperscript{\textnormal{2}}\quad
  Shiyuan Zhang\textsuperscript{\textnormal{1}}\quad
  Shichang Zhang\textsuperscript{\textnormal{3}}\quad
  Himabindu Lakkaraju\textsuperscript{\textnormal{3}}\AND
  Ruijiang Gao\textsuperscript{\textnormal{4}}\quad
  Chris Donahue\textsuperscript{\textnormal{5}}\quad
  Jiaqi W.\ Ma\textsuperscript{\textnormal{1}}\\
  \\
  \textsuperscript{\textnormal{1}}University of Illinois Urbana-Champaign\quad
  \textsuperscript{\textnormal{2}}University of Michigan\\
  \textsuperscript{\textnormal{3}}Harvard University\\
  \textsuperscript{\textnormal{4}}The University of Texas at Dallas\\
  \textsuperscript{\textnormal{5}}Carnegie Mellon University\\
}

\renewcommand{\headeright}{Working Paper}
\renewcommand{\undertitle}{Working Paper}
\renewcommand{\shorttitle}{Computational Copyright}
\newtheorem{lemma}{Lemma}[section]

\hypersetup{
pdftitle={Computational Copyright: Towards A Royalty Model for Music Generative AI},
pdfsubject={cs.AI, cs.LG},
pdfauthor={Junwei~Deng, Jiaqi~Ma},
pdfkeywords={Generative AI, Copyright, Music Royalty, Data Attribution},
}

\begin{document}
\maketitle

\begin{abstract}
The rapid rise of generative AI has intensified copyright and economic tensions in creative industries, particularly in music. Current approaches addressing this challenge often focus on preventing infringement or establishing one-time licensing, which fail to provide the sustainable, recurring economic incentives necessary to maintain creative ecosystems. To address this gap, we propose \emph{Generative Content ID}, a framework for scalable and faithful royalty attribution in music generative AI. Adapting the idea of YouTube’s Content ID, it attributes the value of AI-generated music back to the specific training content that causally influenced its generation, a process we term as \emph{causal attribution}. However, naively quantifying the causal influence requires counterfactually retraining the model on subsets of training data, which is infeasible. We address this challenge using efficient Training Data Attribution (TDA) methods to approximate causal attribution at scale.

We further conduct empirical analysis of the framework on public and proprietary datasets. First, we demonstrate that the scalable TDA methods provide a faithful approximation of the ``gold-standard'' but costly retraining-based causal attribution, showing the feasibility of the proposed royalty framework. Second, we investigate the relationship between the \emph{perceived similarity} employed by legal practices and our \emph{causal attribution} reflecting the true AI training mechanics. We find that while perceived similarity can capture the most influential samples, it fails to account for the broader data contribution that drives model utility, suggesting similarity-based legal proxies are ill-suited for royalty distribution. Finally, our economic simulations reveal that the specific design of attribution mechanisms significantly impacts income distribution, acting as a tunable lever that platforms can use to balance between highly concentrated and more equitable economies.

Overall, this work provides a principled and operational foundation for royalty-based economic governance of music generative AI, offering actionable insights for researchers, policymakers, and industry stakeholders seeking pathways for a creative ecosystem with sustainable collaborations between human and AI.
\end{abstract}

\keywords{Generative AI \and Copyright \and Music Royalty \and Data Attribution}

\section{Introduction}\label{sec:intro}

Recent advancements in generative AI have significantly impacted creative industries, leading to a surge in AI-generated content across images~\citep{zhou2024generative}, music~\citep{dhariwal2020jukebox}, text~\citep{doshi2024generative}, and software~\citep{10.1145/3747588}. This rapid evolution has raised complex legal challenges, especially concerning copyright issues~\citep{henderson2023foundation,samuelson2023generative,sag2023copyright,franceschelli2022copyright}. Copyright laws cover a range of rights, including protection of original works, controlling their reproduction, and managing the distribution of profits from these works. The emergence of generative AI poses multifaceted challenges in this regard, as it blurs the lines of authorship and originality. Compounding these challenges, some studies show that most copyright holders are neither compensated nor aware of how their work is used to train generative models~\citep{morreale2023data, kyi2025governance}. These concerns are also highlighted by high-profile lawsuits across media domains,
from text in \citet{NYT_Microsoft_OpenAI} to music in \citet{UMG_v_Suno,UMG_v_Udio} and \citet{Concord_Music_Group_v_Anthropic}.

Economic considerations are at the heart of these copyright disputes. The U.S. Constitution frames copyright as an incentive mechanism to ``promote the Progress of Science and useful Arts'' by granting creators exclusive rights\footnote{Article 1, Section 8, Clause 8 of the U.S. Constitution.}. It is well-established that uncompensated distribution of creative work decreases investment in creation~\citep{danaher2020piracy}. In music market, \citet{bhattacharjee2007effect} found that piracy has a negative impact on low-ranked albums. Consequently, courts have emphasized that economic compensation is vital for stimulating production~\citep{mazer1954, eldred2003}. In the modern digital music market, this principle is operationalized through royalty and licensing frameworks. Platforms like Spotify and YouTube utilize systems like \emph{Streamshare} and \emph{Content ID} to distribute recurring revenue to rights holders based on consumption~\citep{aguiar2018streaming}. Royalty models are generally preferred over one-time licensing because they align the interests of the creator with the platform: as the platform grows, the creator's revenue grows, ensuring sustainable incentives for future creation.

The urgency of implementing such economic principles in the AI era was starkly highlighted in the recent ruling of \citet{Bartz_v_Anthropic}. Centered on the use of pirated books for model training, the case resulted in a \$1.5 billion settlement—reportedly the largest payout in U.S. copyright history~\citep{Metz_Anthropic_2025}. This landmark event underscores the massive liability AI developers face and signals that the current "wild west" approach to training data is legally and financially unsustainable.

Broadly, solutions to this crisis fall into two categories: \emph{preventive solutions} and \emph{economic solutions}. Preventive solutions focus on technically constraining the model to stop it from regurgitating training data or generating content that resembles copyrighted works~\citep{vyas2023provable,chu2023protect,li2023mitigate}. However, this direction faces significant limitations. First, preventive guardrails are often brittle and vulnerable to adversarial jailbreaking attacks. Second, aggressively filtering training data or constraining generation often degrades the general quality and utility of the model. Most importantly, preventive measures defeat the economic purpose of copyright: they aim to exclude creators from the AI ecosystem rather than compensating them for their contribution to it.

Therefore, we advocate for \emph{economic solutions}, specifically those based on royalties. While one-time licensing (e.g., ``Fairly Trained'' certifications\footnote{\url{https://www.fairlytrained.org}.}) offers a technically simpler stopgap, it fails to provide the recurring, proportional compensation that has stabilized the digital music industry. A more equitable solution may involve a royalty-based framework where creators are compensated based on how their data contributes to generated outputs. However, implementing this for generative AI presents a fundamental technical challenge: \emph{attribution}. Unlike streaming, where a specific song is played, generative AI consumes training data to synthesize new content. We must determine how to attribute the ``consumption'' of a generated song back to the specific training tracks that influenced its creation.

To address this, we propose \emph{Generative Content ID}, a framework for scalable and faithful royalty distribution in music generative AI. Drawing on the logic of YouTube's Content ID, which measures the indirect consumption of music inside a video and redistributes revenue accordingly, our framework measures the indirect consumption of training data. Specifically, we attribute the value of a generated music piece back to the training samples that causally influenced its generation. Establishing this link requires a rigorous definition of influence. We advocate for \emph{causal attribution}: estimating how the model's generation would change if a specific training sample had been removed (counterfactual prediction). While the ``gold standard'' for this is retraining the model on subsets of training data (which is economically infeasible), we address this scalability challenge by leveraging efficient Training Data Attribution (TDA) methods~\citep{koh2017understanding,deng2025survey}, which approximately quantify the counterfactual prediction without expensive retraining.

In this study, we implement Generative Content ID and investigate three key research questions: 
\begin{enumerate}
    \item \textbf{Validity:} Does the practical implementation of the framework via TDA accurately approximate the ideal, retraining-based causal attribution?
    \item \textbf{Comparison with Legal Proxies:} Existing legal practice often relies on \emph{perceived similarity} (how much one piece of music resembles the other) as a proxy for infringement. We investigate the relationship between this legal proxy (\emph{perceived attribution}) and the technical reality of model influence (\emph{causal attribution}). By empirically measuring the alignment between these two metrics, we aim to provide data-driven insights into how well current similarity-based standards capture the actual mechanics of generative AI training.
    \item \textbf{Economic Implications:} What are the economic implications of deploying Generative Content ID regarding income distribution among creators?
\end{enumerate}

Our experiments using public and proprietary music industry data yield three key findings. First, the proposed Generative Content ID, implemented via gradient-based TDA, achieves reasonably high correlation with the retraining ``gold standard,'' demonstrating that scalable causal attribution is feasible. Second, we find that while perceived attribution matches causal attribution for the most heavily influential training pieces, the overall alignment is poor. This suggests that perceived attribution may catch the most salient ``memorization'' cases, it fails to capture the broader, subtle causal usage of data that drives model performance, and is therefore not well-suited as a basis for royalty distribution. More broadly, this result also provides empirical insights regarding legal standards of copyright infringement in the AI era. Finally, our economic simulations reveal that the specific mechanism of royalty distribution acts as a tunable knob, significantly impacting income inequality. By adjusting these mechanisms, platforms can strategically decide between highly concentrated and more equitable economies.

The remainder of this paper is organized as follows. We begin in Section~\ref{sec:case} by reviewing the background of digital copyright and analyzing existing royalty frameworks, utilizing Spotify and YouTube as case studies to identify key design requirements. Section~\ref{sec:attribution} establishes a conceptual royalty framework for AI music generation and proposes our core technical mechanism: Generative Content ID. In Section~\ref{sec:experiment}, we present our empirical analysis, which evaluates the validity of the framework, compares the technical reality of causal attribution against the legal proxy of perceived attribution, and simulates the economic implications of various distribution mechanisms. Finally, we conclude in Section~\ref{sec:conclusion} with a discussion of the broader managerial and policy implications of this work.
\section{Background}\label{sec:case}

In this section, we provide the necessary context for our proposed framework. We first review the legal standards and economic incentives that have historically shaped digital copyright, establishing the precedent for moving from ``takedown'' models to ``licensing'' models. We then examine existing royalty frameworks in the digital music industry through the lens of two representative case studies, Spotify and YouTube, to identify the design requirements for a generative AI royalty system.

\subsection{Legal Standards and Economic Incentives in Copyright}

\vpara{Legal Standards.} Legal issues related to generative AI copyright have attracted considerable interest. One of the primary discussions is the applicability of the \emph{fair use} doctrine\footnote{See Section 107 of the Copyright Act: \url{https://www.copyright.gov/title17/92chap1.html\#107}.}. While previous studies~\citep{sag2018new} and lawsuits\footnote{Authors Guild v. Google, Inc., 804 F.3d 202 (2d Cir. 2015)} established fair use in text data mining, recent scholarship questions its suitability for AI-generated content~\citep{henderson2023foundation,samuelson2023generative,sag2023copyright,franceschelli2022copyright,peukert2024economics}. One of the main reasons for this shift is that AI-generated content can be substantially similar to original works in the training data, and directly impact the market value of the original, the latter of which is a key factor in fair use analysis~\citep{henderson2023foundation,peukert2024economics}.

Furthermore, the legal standard for determining copyright infringement for AI generation remains legally complex~\citep{lee2023talkin}. Under U.S. copyright law, derivative works such as AI outputs may infringe on the copyright holder’s adaptation rights\footnote{See Section 106 of the Copyright Act: \url{https://www.copyright.gov/title17/92chap1.html\#106}.} if certain doctrinal tests are satisfied. Traditional infringement analysis typically requires establishing two elements: access and similarity~\citep{vyas2023provable,lee2023talkin}. Access considers whether any part of an AI generation was directly influenced by a specific training example, and similarity considers if the generation is substantially similar to a training example. Existing legal practice often relies on the similarity (perceived attribution) as a proxy for the infringement test. However, the legal community has recognized a technical gap in definitively determining which training samples are accessed by the generation~\citep{lim2023generative,lee2023talkin,gans2024copyright}. The solution developed in this paper can be viewed as an attempt towards bridging this gap by providing a technical means to verify access and influence.

\vpara{Economic Incentives.} Beyond legal definitions, the economic mechanisms of copyright enforcement have evolved significantly with digitalization of music~\citep{peukert2024economics,valdovinos2021you,gans2024copyright}. Historically, the dominant regime for handling derivative works on the internet was \emph{Notice and Takedown} (N\&T)\footnote{See \url{https://www.copyright.gov/dmca/}.}, employed by platforms like YouTube~\citep{VanDerSar2021YouTube} and TikTok~\citep{valdovinos2021you}. However, recent studies suggest that N\&T systems can be economically inefficient~\citep{peukert2024economics}. Specifically, copyright holders, especially in the music industry, could benefit from derivative works like user-generated videos~\citep{gans2015remix}. Forced takedowns eliminate the derivative work, erasing potential value for both creators of the derivative work and copyright holders of the original~\citep{peukert2024economics}.

Consequently, there is a marked transition from N\&T systems toward \emph{algorithmic licensing}~\citep{peukert2024economics}. Algorithmic licensing creates a market where copyright owners can monetize, rather than remove, derivative works. For instance, YouTube's Content ID allows rights holders to claim revenue from user-generated videos utilizing their music. Empirical evidence indicates that when given the choice, rights holders mostly prefer monetization over takedowns~\citep{copyright21for2021}. This historical shift suggests that the sustainable solution for Generative AI is not to ban training data (N\&T), but to create a mechanism for algorithmic licensing and revenue sharing.

\subsection{Digital Music Royalty Frameworks}

In this section, we examine the specific royalty mechanisms that facilitate the algorithmic licensing described above\footnote{Please refer to Appendix~\ref{concept-of-music-royalties} for a more detailed discussion regarding music royalties that are prevalent in the industry.}. While one-time licensing fees exist, we focus on consumption-based compensation models, as they have demonstrated long-term viability in music streaming by ensuring that copyright holder revenue grows proportionally with platform success~\citep{aguiar2018streaming}.

There are some common revenue distribution and licensing mechanisms on modern consumption-based digital platforms:
\begin{itemize}
    \item \emph{Pro-Rata Model / User-Centric Model}: These models govern how a centralized revenue pool (e.g., subscriptions) is divided among creators. In the pro-rata model, platforms pool all subscription revenue and distribute it based on each copyright owner's share of total streams during a given period. Some representative platforms are Spotify and Apple Music. In the user-centric model, the subscriber's fees are distributed solely to the specific artists they stream. Some representative platforms are SoundCloud and Deezer. The main difference between the two models is how the total revenue is aggregated for each artist. 
    \item \emph{Micro-Sync Licensing Model}: This model treats each use of music in user-generated content as a micro-synchronization license, where royalties are generated when a song is paired with the user-generated content, such as videos in YouTube. Payments are typically tied to the revenue from the specific videos that feature the music. In practice, most companies (e.g., Youtube, TikTok) using the micro-sync model adopt the pro-rata model for revenue distribution. 
\end{itemize}

\paragraph{\textbf{Related Literature on Existing Royalty Allocation Mechanisms:}} Prior research in management science and information systems has extensively examined the economic implications of revenue allocation mechanisms on digital platforms. \citet{alaei2022revenue} provide a critical comparative analysis of pro-rata versus user-centric allocation strategies, demonstrating that despite the cross-subsidization inherent in pro-rata models, they may yield higher payments for artists with high-consumption user bases and can be preferred by platforms under specific conditions. Complementing this, \citet{jain2021compensating} analyze how producer competition and customer base size affect compensation equilibrium, finding that increased competition can paradoxically lead to higher content quality and producer profits. From an axiomatic perspective, \citet{bergantinos2020sharing} explore revenue sharing in broadcasting, characterizing rules that balance fairness and operational constraints. Building on this axiomatic approach, \citet{bergantinos2025revenue} specifically address the music streaming industry, providing game-theoretical foundations for the pro-rata and user-centric methods and characterizing a family of ``weighted'' indices that compromise between the two. 
Furthermore, the challenge of attributing value to specific contributions has been explored by \citet{singal2022shapley} in online advertising, who developed axiomatic frameworks for attribution based on Shapley values for user journeys. Our work extends these foundational inquiries into the novel domain of generative AI, where the ``contribution'' is not a direct view or click, but latent training data influence. Importantly, our proposed attribution framework adapts the micro-sync licensing framework for modern generative AI and is compatible with both pro-rata and user-centric models in terms of revenue distribution. We adopt the pro-rata model for the economic analysis in the remainder of this paper due to its prevalence in the current industry landscape.

\subsection{Case Studies: Spotify and YouTube}

To design a royalty framework for AI music generation platforms, we analyze two major digital media platforms: Spotify and YouTube Video. Spotify is the largest music streaming platform in the world and a representative of \emph{pro-rata model}. YouTube is a major video and media sharing platform. Though featured with multifaceted compensating mechanisms, YouTube is a good representative of \emph{micro-sync model}. 
We select these two platforms because they serve as the clearest archetypes for the dominant revenue distribution paradigms. While other major players like Apple Music, Amazon Music, and TikTok operate with their own specific nuances, their underlying economic structures largely align with the logic exemplified by Spotify and YouTube, respectively.
Both platforms have a significant amount of music content and generate revenue through multiple sources. We review them by addressing the following key questions: 1) Who are the stakeholders? 2) What are the sources of revenue? 3) How to determine the royalty distribution for revenue sharing?

\subsubsection{Spotify's Royalty Framework}\label{sec:spotify}
\begin{figure}[h]
    \centering
    \includegraphics[width=0.7\linewidth]{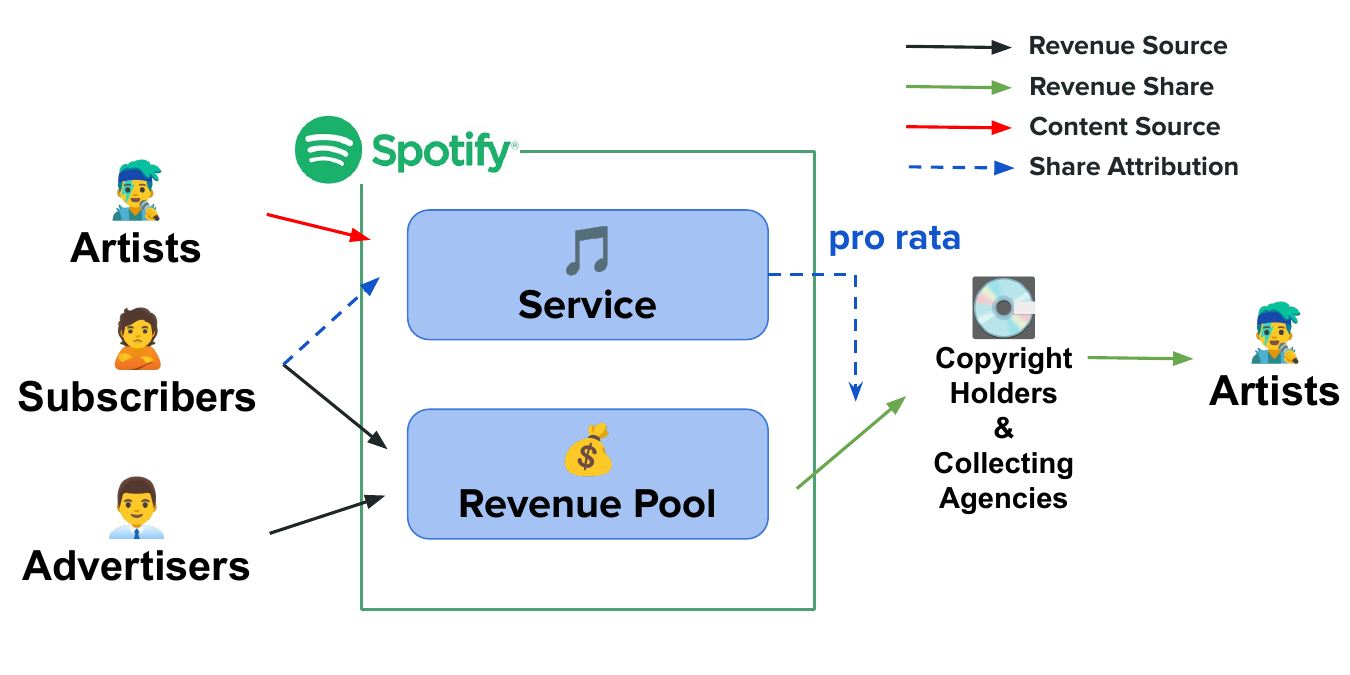}
    \caption{The royalty framework of Spotify.}
    \label{fig:spotify-diagram}
\end{figure}

Spotify employs a centralized method for sharing its revenue with copyright owners, primarily via \emph{pro-rata model}. The process involves determining Spotify's total revenue from various sources and subsequently calculating the royalty distribution for copyright owners. The process is shown in Figure~\ref{fig:spotify-diagram}.

\vpara{Stakeholders.} Spotify's royalty framework involves several key groups of stakeholders, in addition to the streaming platform itself. These groups\footnote{Please refer to Appendix~\ref{appendix:detailed-stakeholder-description} for detailed description of these groups of stakeholders.} are (1) \emph{artists and creators}, (2) \emph{record labels and music publishers}, (3) \emph{music rights societies and collecting agencies}, (4) \emph{listeners and subscribers}, and (5) \emph{advertisers}.

Stakeholders in groups 1, 2, and 3 receive revenue shares from Spotify, while groups 4 and 5 contribute to the generation of Spotify's revenue. Typically, Spotify directly interacts with stakeholders in groups 2 and 3, which we denote as ``Copyright Holders \& Collecting Agencies'' in Figure~\ref{fig:spotify-diagram}. Individual artists and creators often have contracts with these labels, publishers, or music rights agencies, and do not directly engage with Spotify in the financial aspect of their music streaming.

\vpara{Revenue Sources.} The major revenue sources of Spotify can be divided into two categories: subscription and advertisement. In 2021, premium subscriptions accounted for 88\% of Spotify's revenue while advertisements accounted for the remaining 12\%~\citep{johnstone2023spotify}. The two revenue sources lead to the formation of separate revenue pools, which are also calculated separately for different countries or regions.

\vpara{Royalty Distribution.} In the pro-rata model, the royalty for a track is calculated by applying its percentage of total platform streams to the revenue pool. This method ensures that royalty distribution is directly proportional to the consumption frequency on the platform.

\subsubsection{YouTube Video's Royalty Framework}\label{sec:case-study-youtube}

\begin{figure}[h]
    \centering
    \includegraphics[width=0.7\linewidth]{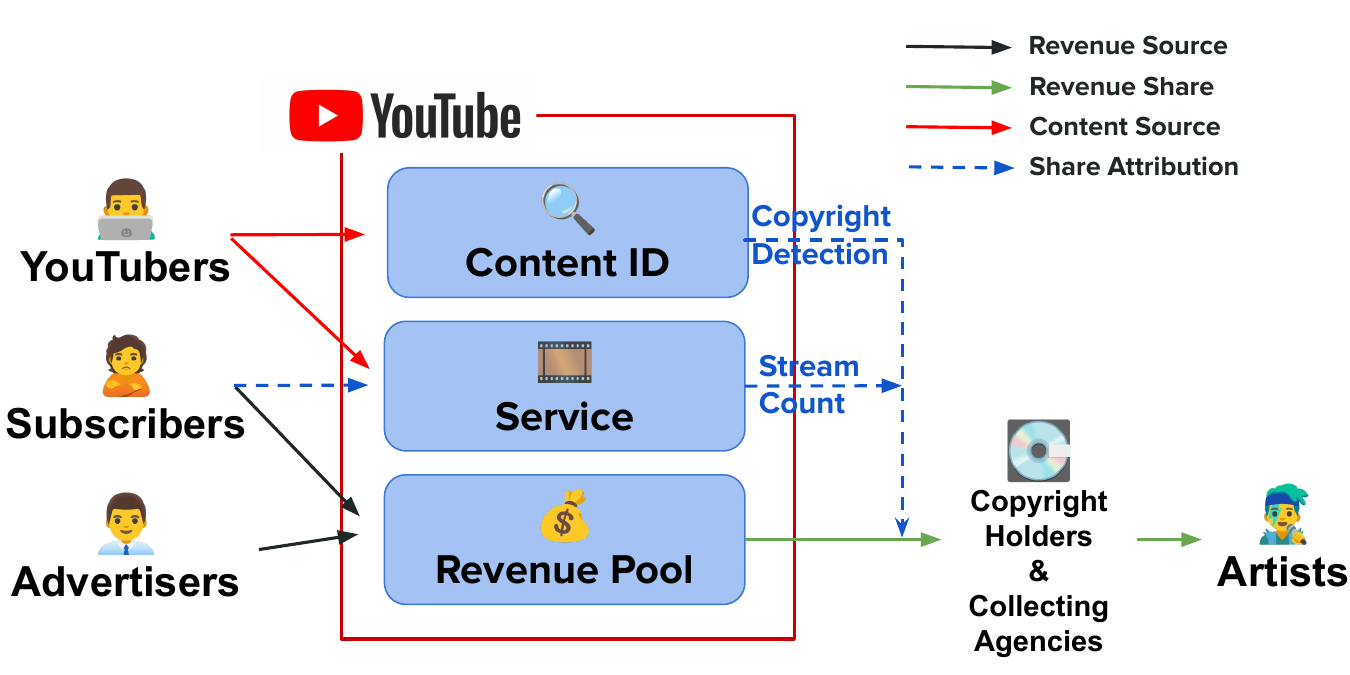}
    \caption{The royalty framework of Youtube Video.}
    \label{fig:youtube-diagram}
\end{figure}

YouTube Video's model for compensating music copyright owners is multifaceted, offering various methods for monetizing the content. (1) \emph{YouTube Partner Program:} Music copyright owners can join the YouTube Partner Program, uploading music (videos) to their official channels. Revenue is shared based on user views of their content. (2) \emph{Content ID and Royalty:} Owners can earn from videos using their music through the Content ID system. This system uses fingerprinting and machine learning to identify copyrighted content in uploaded videos and allocates revenue from these videos to the copyright owners. (3) \emph{One-time Licensing:} Owners can also license their music directly to a YouTube video for a one-time payment. The process using Content ID and royalty-based framework is shown in Figure~\ref{fig:youtube-diagram}.

\vpara{Stakeholders.} The stakeholders involved in the first monetization method above are similar to those in Spotify's royalty framework. However, the second and third methods introduce additional parties: video creators and third-party licensing platforms. Video creators (YouTubers) are the owners of YouTube videos that can be viewed as derivative content of copyrighted music. Third-party licensing platforms help video creators obtain licenses for music used in their videos. These platforms often have direct licensing agreements with YouTube and music rights owners, offering a streamlined process for video creators to legally use music in their videos. 

\vpara{Revenue Sources.} For the first two methods, royalties come from YouTube's revenue streams. YouTube generates revenue primarily through advertisements and, to a lesser extent, through premium subscriptions. 

\vpara{Royalty Distribution.} A central challenge in YouTube's royalty framework is identifying copyrighted elements in the user-generated videos and attributing them to the copyright holders. Unlike Spotify, where the platform hosts the file, YouTube users upload content where copyright provenance is difficult to trace manually. To address this, YouTube developed \emph{Content ID}, an algorithmic solution that uses fingerprinting to identify copyrighted music within uploads. When a match is found, the system attributes the video's revenue to the music copyright owner. Although Content ID faces criticism for false positives~\citep{VanDerSar2021YouTube,mckay2011youtube}, it remains the industry standard for identifying derivative usage. 

The Content ID system serves as the technical foundation that enables the second and third methods of YouTube's revenue sharing. In the second method, it establishes the indirect consumption measure of the copyrighted music in videos and allocates revenue to the copyright holders. In the third method, while synchronization licenses might be obtained through third-party licensing platforms outside of YouTube, the presence of the Content ID system encourages them to secure these licenses. 

\section{A Computational Royalty Framework for Music Generative AI}\label{sec:attribution}

In this section, we propose a computational royalty framework for music generative AI. We begin by analyzing the emerging business models of AI music to identify design requirements. We then conceptualize the royalty framework by drawing a structural analogy to YouTube's ecosystem. Crucially, we identify the measurement of \emph{indirect consumption}---how training data influences generation---as the critical technical bottleneck preventing the deployment of such a framework. To bridge this gap, we introduce \emph{Generative Content ID}, a causal attribution mechanism designed to trace generated content back to training data. Finally, we contrast this approach with a naive alternative approach based on perceived similarity.

\subsection{Business Models of AI Music Generation Platforms}

While the landscape of AI music generation platforms is still rapidly evolving, there have been a few common business models emerging~\citep{suno2025best}. We summarize these business models in terms of stakeholders, revenue sources and services.

\vpara{Services.} The backbone of AI music generation platforms is the generative AI model trained on a large corpus of existing music, which often includes copyrighted music. Platforms leverage these models to offer a variety of services, with some examples including (1) \emph{creation tools} assisting both amateur and professional users in music composition, such as Udio~\citep{Udio2025}\footnote{The examples provided may offer more than just a single type of service.}; (2) \emph{AI music streaming} with libraries of AI-generated music, such as Suno~\citep{sunoai2023}; (3) \emph{custom music solutions} for corporate clients, such as loudly~\citep{Loudly2025}; and (4) \emph{educational tools} for interactive music learning experiences such as Riyaz~\citep{Riyaz2025}.

\vpara{Stakeholders.} The potential stakeholders involved in AI music generation platforms have significant overlaps with those on traditional music platforms, as summarized in the five groups in Section~\ref{sec:spotify}. The groups 1, 2, and 3, including \emph{artists} and \emph{record labels}, remain the primary recipients of revenue sharing. Similarly, \emph{end users} and \emph{advertisers}, corresponding to groups 4 and 5, contribute to the platforms' revenue. A new stakeholder group includes \emph{corporate clients} (e.g., game developers and filmmakers) who require bespoke, royalty-free generative solutions. 

\vpara{Revenue Sources.} Platforms typically aggregate revenue through four primary channels~\citep{pateriya2025monetizing,monetizely2025genai}: (1) \emph{subscription fees} for access to music creation tools, libraries, or streaming services; (2) \emph{licensing fees} for downstream commercial use of generated music, e.g., in a YouTube video or a video game; (3) \emph{advertisements} on free-tier streaming services; and (4) \emph{custom composition fees} for corporate clients on bespoke music creation services.

\subsection{Proposed Royalty Framework for Music Generative AI}\label{sec:model}

\begin{figure}[h]
    \centering
    \includegraphics[width=0.7\linewidth]{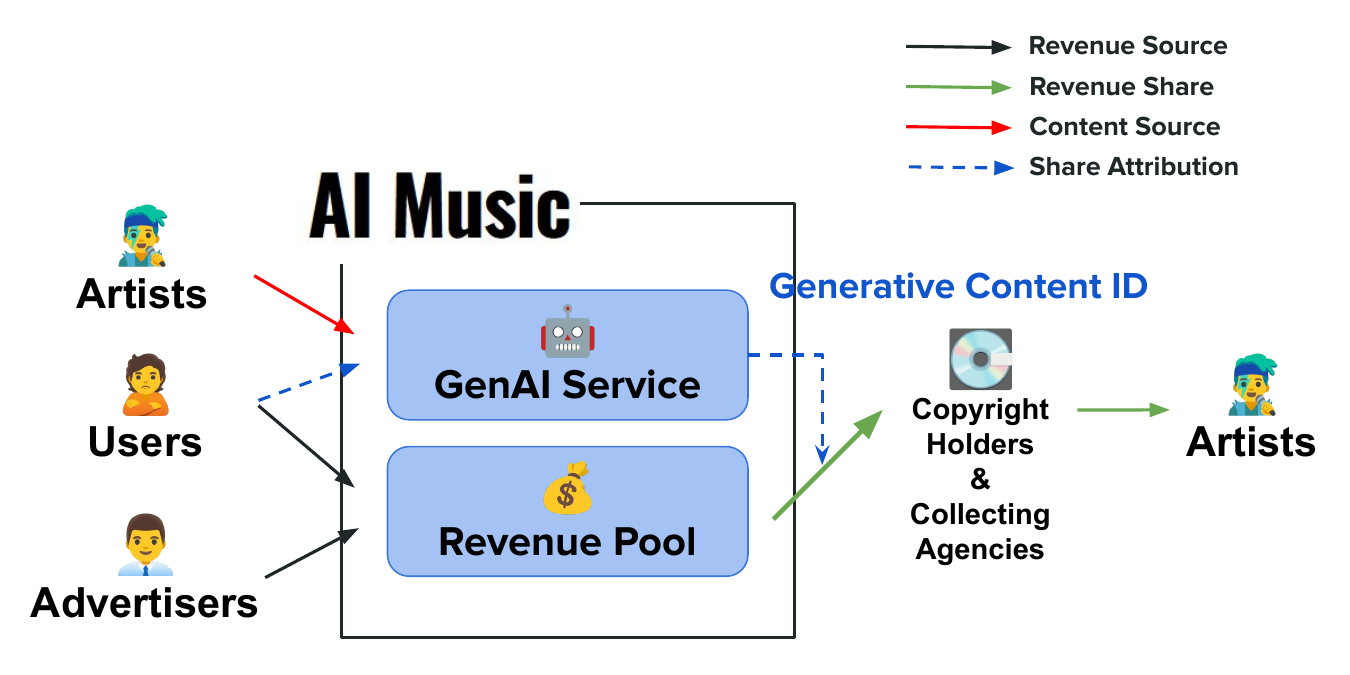}
    \caption{The computational royalty framework designed for music generative AI.}
    \label{fig:music-ai-diagram}
\end{figure}

Given the close alignment of the stakeholders and revenue sources between the AI music generation platforms and traditional music platforms, it is natural to consider adapting existing royalty frameworks, such as those used by Spotify or YouTube, to the design for music generative AI. In particular, the business models in AI platforms exhibit a similar pattern observed in our case study, where revenue flows through the platform rather than being tied to individual pieces of music. 

We draw a structural analogy to YouTube's model. As illustrated in Figure~\ref{fig:music-ai-diagram}, we propose a royalty framework that consists of three key elements:
\begin{enumerate}
    \item \emph{Formation of Revenue Pools.} Similar to Spotify and YouTube, the AI platform aggregates revenue from a variety of revenue sources (subscriptions, advertisements, licensing fees, etc.), forming revenue pools.
    \item \emph{Generation and Consumption.} Platform users (creators, streaming listeners, or corporate clients) generate and/or consume new music by interacting with the AI services.
    \item \emph{Revenue Distribution Based on Indirect Consumption.} Revenue is distributed to rights holders based on the \emph{indirect consumption} of their work. This is calculated by determining how much their copyrighted training data contributed to the generated content, scaled by the revenue that generated content brings to the platform.
\end{enumerate}

\paragraph{\textbf{The Technical Bottleneck in Measuring Consumption:}} A critical technical challenge for realizing this royalty framework lies in the third element, which requires measuring the indirect consumption of copyrighted training content through the AI services. In the YouTube ecosystem, the contribution of a piece of copyrighted music is explicit: a user uploads a video containing this specific music. YouTube's Content ID detects direct matches of music in videos and measures the contribution by the length of time the music is being played in the video. In music generative AI, however, the contribution is latent; the model learns statistical patterns rather than storing exact copies, making it unclear how to measure the contribution of a specific training sample to a specific generation.

Nevertheless, once this technical challenge is resolved, the indirect consumption of copyrighted training content can be quantified in a manner structurally analogous to YouTube’s model. Specifically, the contribution of each training sample to a generated piece can be multiplied by the consumption of that generated piece, such as stream counts or downstream licenses it receives. This restores a ``pay-per-play'' logic: rights holders are compensated proportionally to how much their works causally contribute to the AI-generated outputs that users ultimately consume.

\subsection{Generative Content ID}\label{sec:generative-content-ID}

To address the technical challenge identified in Section~\ref{sec:model}, we introduce \emph{Generative Content ID}, a framework spiritually motivated by YouTube's Content ID, but designed specifically to measure the indirect consumption in music generative AI.

\subsubsection{Tracing Generated Music to Copyrighted Training Content}

The goal of the framework is to trace the content being directly consumed, i.e., the generated music, back to the copyrighted music contributed to their generation. This is analogous to how YouTube's Content ID trace a YouTube video back to the copyrighted music contained in the video. We establish this tracing mechanism through \emph{causal attribution}, which frames the task as a counterfactual prediction problem: how does the removal of a particular piece of training music from the training dataset affect the output of the model retrained on the remaining dataset? This change in output serves as a measure of the removed data point's influence, i.e., \emph{leave-one-out} influence, on the specific model output~\citep{koh2017understanding}.

Now we formalize the definition of leave-one-out influence. For any piece of music segment $m$ and any music generative model $h$, we first define a utility function $f$ that maps $m$ and $h$ to a real value $f(m, h)$, which quantifies how well $h$ generates $m$. We let $h_{S}$ be a model retrained on the dataset of $n$ pieces of music $S = \{m_1, \dots, m_n\}$. The \emph{attribution score} $I(m_i, \hat{m})$, representing the leave-one-out influence of a training piece $m_i\in S$ on a new piece of $h$-generated music $\hat{m}$, is defined as 
\begin{align}
    I(m_i, \hat{m}) = f(\hat{m}, h_{S}) - f(\hat{m}, h_{S\setminus\{m_i\}}), \label{eq:influence}
\end{align}

In practice, the utility function $f(\hat{m}, h)$ can be defined as the conditional log-likelihood\footnote{Negative log-likelihood is often used as the loss function for the generative models.} of the generated music segment $\hat{m}$ given a user-provided prompt and $h$. A large score $I(m_i, \hat{m})$ then signifies that removing the training piece $m_i$ decreased the log-likelihood of generating $\hat{m}$, thereby implying that $m_i$ was useful to the generation process.

\subsubsection{Music Generative AI: Model and Utility Function}

Next, we discuss concrete choices of the utility function $f$ for the influence tracing in music generative AI.

\vpara{Formal Definition of AI Music Generation.} We first introduce the formal definition of AI music generation. Broadly, there are two major paradigms of AI music generation: \emph{waveform music generation}~\citep{oord2016wavenet} and \emph{symbolic music generation}~\citep{pmlr-v70-hadjeres17a}. Waveform music generation involves the direct synthesis of a music's waveform, with examples including WaveNet~\citep{oord2016wavenet} and Stable Audio Open~\citep{evans2024stableaudioopen}. Symbolic music generation involves creating music in a symbolic format, such as the Musical Instrument Digital Interface (MIDI) format~\citep{de2017understanding}. Some representative methods include Music Transformer~\citep{huang2018music} and DeepBach~\citep{pmlr-v70-hadjeres17a}. 

Both paradigms are used in modern AI music-generation platforms. Although commercial platforms rarely disclose full architectural details, we can infer aspects of their generation pipelines from public information. Platforms such as AIVA~\citep{aiva2016} and Musia~\citep{musia2024} appear to rely more on symbolic modeling, while platforms like Suno AI~\citep{sunoai2023} and Stable Audio~\citep{stableaudio2023} emphasize waveform-level generation. A growing industry trend is hybrid modeling, combining the creative richness of waveform synthesis with the editability and structure provided by symbolic representations~\citep{suno2025best}.

As a proof of concept, we focus on symbolic music generation in this study, where a piece of music $m$ is represented as a sequence of $k$ discrete events $m =(e_1, e_2, \ldots, e_k)$ with each $e_k \in \mathcal{V}$ from a vocabulary of musical events (e.g., notes, durations, instruments). A symbolic music generation model, denoted as $h_S$, where $S = \{m_1, \ldots, m_n\}$ is training dataset, often takes the form of as an \emph{autoregressive generative model}, such as Music Transformer~\citep{huang2018music}. Such model $h_S$ takes in a prompt music segment with $p$ events, $(e_{k-p}, e_{k-p+1}, \ldots, e_{k-1})$, and calculates the conditional probability distribution of the next event, $P(e_k \mid e_{k-p}, e_{k-p+1}, \ldots, e_{k-1}; h_S)$. The generated music can be defined as $\hat{m}=(\hat{e}_1, \hat{e}_2, \ldots, \hat{e}_l)$ with each $\hat{e}_l \in \mathcal{V}$. A symbolic music can be generated by recursively sampling from this conditional probability and updating the prompt with moving window.
This formulation is widely used in symbolic music AI systems~\citep{huang2018music, openai2019musenet}, and we ground our subsequent discussion of the attribution problem with this model formulation.

\vpara{Utility Function of Music Generative AI.} In the context of AI music generation, we can define the influence of a piece of training music on a piece of generated music in terms of the change in the likelihood of the model producing that generated music. Specifically, the utility function $f$ can be defined as the (log-)likelihood of the music segment $m$ being generated by model $h$. In this case, $I(m_i, \hat{m})$ measures the change of likelihood for $\hat{m}$ being generated when $m_i$ is removed from the training corpus.

\vpara{Two Levels of Attribution.} 
For music generative AI, the users can use a part of the generated segment with arbitrary number of events for monetization purposes, thus it is important for us to quantify the influence of training data on generated music segments of varying lengths.
We can define two instances of the data attribution problem, respectively \emph{event-level attribution} and \emph{segment-level attribution}. The event-level attribution corresponds to a special case where $\hat{m}$ has a single event, i.e., $|\hat{m}| = 1$. The segment-level attribution corresponds to the general case where $\hat{m}$ has multiple events, i.e., $|\hat{m}| > 1$. The two instances provide different granularity of attribution scores. In an autoregressive symbolic music generation model, the music is generated event by event. Therefore, the same training data point could have different influences when generating different events in a segment. The event-level attribution provides a way to capture this nuance. On the other hand, the segment-level attribution looks at the influence of training data on a larger scale, focusing on the overall structure and composition of a generated music segment.

Since generating one event can be viewed as a classification problem, we can apply existing data attribution methods for classification models to event-level attribution. We then extend this logic to segment-level attribution through an additive framework: when the utility function is defined as the log-likelihood, 
i.e., $f(\hat{m}, h) = \log P(\hat{m}; h)$. Let $\hat{m}=(e_{k}, \ldots, e_l)$ for some $k < l$. We have the following Lemma. 

\begin{lemma}[Additivity of Influence under Log-Likelihood]
Let $h_S$ be an autoregressive generative model trained on dataset $S$, and let $m_i \in S$ be a specific training sample. Let $\hat{m} = (e_k, \dots, e_l)$ denote a generated sequence of events. If the utility function $f(\hat{m}, h)$ is defined as the log-likelihood, i.e., $f(\hat{m}, h) = \log P(\hat{m}; h)$, then the segment-level influence $I(m_i, \hat{m})$ decomposes additively into the sum of event-level influences:
\begin{equation}
    I(m_i, \hat{m}) = \sum_{j=k}^l I(m_i, \{e_j\}).
\end{equation}
\end{lemma}

\begin{proof}
Since the generation of events follows an autoregressive process, we apply the chain rule of probability to the utility function $f(\hat{m}, h_S)$:
\begin{align*}
    f(\hat{m}, h_S) 
    &= \log P(\hat{m}; h_S) \\
    &= \sum_{j=k}^l \log P(e_j \mid e_k, \dots, e_{j-1}; h_S) \\
    &= \sum_{j=k}^l f(\{e_j\}, h_S).
\end{align*}
By definition, the influence of a training sample $m_i$ on the generated segment $\hat{m}$ is the difference in utility when the model is retrained on $S \setminus \{m_i\}$. Substituting the additive decomposition derived above:
\begin{align*}
    I(m_i, \hat{m}) 
    &= f(\hat{m}, h_S) - f(\hat{m}, h_{S \setminus \{m_i\}}) \\
    &= \sum_{j=k}^l f(\{e_j\}, h_S) - \sum_{j=k}^l f(\{e_j\}, h_{S \setminus \{m_i\}}) \\
    &= \sum_{j=k}^l \left( f(\{e_j\}, h_S) - f(\{e_j\}, h_{S \setminus \{m_i\}}) \right) \\
    &= \sum_{j=k}^l I(m_i, \{e_j\}).
\end{align*} 
\end{proof}

Thus, segment-level attribution is exactly the sum of the atomic event-level attributions. This decomposition enables the application of attribution techniques designed for classification tasks to the domain of music sequence generation. We also note that this is the first paper applying data attribution methods to music generative models to the best of our knowledge. 

\subsubsection{Scalable Approximation via Efficient Data Attribution Techniques}\label{sssec:scalable-approximation-data-attribution}

Directly calculating $I(m_i, \hat{m})$ requires retraining a model for each training data point $m_i$, which is computationally prohibitive. We design \emph{Generative Content ID} to computationally estimate influence in a scalable way. There has been a rich literature on efficient Training Data Attribution (TDA) methods~\citep{deng2025survey}. Among them, the \emph{influence function}~\citep{koh2017understanding} is a widely used gradient-based method that can approximate causal influence without additional retraining. It estimates the influence by measuring how a small up-weighting of a training example would change the model’s parameters, and the utility function consequently. Specifically, it computes
\begin{equation}
    \hat{I}(m_i, \hat{m}) = - g_{f}(\hat{m})H_S^{-1}g(m_i), \label{eq:influence-function}
\end{equation}
as the approximated causal attribution score for $m_i$ on generating $\hat{m}$. Here $g_{f}(\hat{m}) = \nabla_{\theta} f(\hat{m}, h_S)$ is the gradient of the utility function with respect to the model parameters $\theta$, $H_S^{-1}$ is the inverse Hessian matrix of the training loss $\mathcal{L}(h_S; S) = \frac{1}{n}\sum_{i=1}^n\ell(m_i, h_S)$, and $g(m_i) = \nabla_{\theta} \ell(m_i, h_S)$ is the gradient of the training loss for data point $m_i$. Note that, in comparison to the original leave-one-out influence in Eq.~(\ref{eq:influence}), the influence function in Eq.~(\ref{eq:influence-function}) eliminates the dependency on $h_{S\setminus m_i}$ and can be obtained based on the original model $h_S$, thus eliminating the requirement of retraining. This is a critical advantage, as the brute-force approach of retraining the model for every removed training sample is computationally and economically prohibitive for modern generative models.

In practice, there are several more efficient variants of influence functions that further eliminates the cost of calculating the inverse Hessian $H_S^{-1}$, which can be applied to estimate the attribution scores for large-scale autoregressive symbolic music generation models. 
Avoiding this calculation is essential, as computing and storing the explicit inverse Hessian matrix is often computationally intractable due to the massive parameter space of modern foundational models.
In our empirical analysis, we employ two state-of-the-art methods, \emph{TRAK}~\citep{park2023trak} and \emph{LoGra}~\citep{choe2024your}. Both of the methods reduce the dimension of parameters through random projection. LoGra proposes low-rank projection to further reduce the computational cost. We introduce their technical details in Appendix~\ref{sec:training-data-attribution-algorithms}.
It is worth noting that the proposed royalty framework can work with most of these data attribution methods. This modularity ensures that the framework can adapt to future advancements in efficient AI without requiring changes to the underlying economic logic.

\subsection{An Alternative Solution Through Perceived Attribution}

While the \emph{causal attribution} introduced in Section~\ref{sec:generative-content-ID} represents the technical ground truth of AI training mechanics, it differs from the current legal standard. Existing copyright practice relies on \emph{perceived similarity}, which assesses whether a secondary work is perceived as substantially similar to an original. To understand the relationship between the technical definition and legal standards, we also explore a (naive) solution based on \emph{perceived attribution} as an alternative to the Generative Content ID via \emph{causal attribution}. 

In perceived attribution, we treat the perceived similarity between a training sample $m_i \in S$ and a generated sample $\hat{m}$ as a direct proxy for influence. In legal contexts, such similarity is typically established through human expert analysis spanning multiple aspects, including musical styles (such as genres, rhythms, melodies, or harmonies), technical and instrumental methods (how a musician plays an instrument or sings), or thematic elements (such as themes, messages, or lyrical content)~\citep{begault2013analysis}.

However, relying on human expertise is economically infeasible for auditing large-scale AI systems. Therefore, for our empirical analysis in Section~\ref{ssec:perceived-causal-experiment}, we employ state-of-the-art computational music similarity methods, such as CLAP~\citep{elizalde2023clap}, MERT~\citep{li2024mert}, and PMI~\citep{savage2018quantitative}, as scalable proxies for human judgment. We detail these computational metrics in Appendix~\ref{sec:music-similarity}. As we shall see in Section~\ref{sssec:tda-vs-human-evaluation}, computational music similarity is a good proxy for perceived attribution.

\section{Empirical Analysis}\label{sec:experiment}

In this section, we conduct a systematic empirical analysis of our proposed royalty framework. We orient our analysis around three core research questions:
\begin{itemize}
    \item \emph{Validity}: Does the practical implementation of the Generative Content ID via training data attribution (TDA) accurately approximate the ideal, retraining-based causal attribution? (Section~\ref{ssec:causal-attribution-performance})
    \item \emph{Comparison with Legal Proxies}: How does the technical reality of training data influence (\emph{causal attribution}) compare with current legal proxies based on similarity (\emph{perceived attribution})? (Section~\ref{ssec:perceived-causal-experiment})
    \item \emph{Economic Implications}: What are the economic implications of deploying Generative Content ID regarding income distribution among copyright holders? (Section~\ref{ssec:economic-impact})
\end{itemize}

\subsection{Experiment Settings}

We utilize two distinct model-dataset pairs: 
\begin{enumerate}
    \item \textbf{Music Transformer}~\citep{huang2018music} trained on the \textbf{MAESTRO} dataset~\citep{hawthorne2018enabling}. The dataset comprises approximately 200 hours of MIDI recordings from a piano competition. The music is mostly classical, featuring compositions from the 17th to early 20th century.
    \item \textbf{Anticipatory Music Transformer}~\citep{thickstun2023anticipatory} fine tuned on \textbf{TheoryTab}~\citep{donahue2022melody}, a dataset consisting of unofficial user transcriptions\footnote{TheoryTab: \url{https://www.hooktheory.com/theorytab}} of the melody and harmony of a large number of commercial pop songs. This experimental setting is similar to that of the model powering \emph{Hookpad Aria}~\citep{donahue2024hookpad}, a real-world AI co-pilot tool used by pop songwriters.
\end{enumerate}
Each dataset sampled and divided into training and test sets, where the models are trained on the training sets\footnote{Please refer to Appendix~\ref{more-details-of-experimental-setup} for detailed experimental setup.}. For MAESTRO, we sampled 5,000 pieces of music with 256 events per piece as the training set. For \aria\ , we sampled 28,000 pieces of music with 1,024 events per piece as the training set. From each trained model, we generate a set of music samples (178 for MAESTRO, 500 for \aria\ ) using prompts from their respective test sets. These generated samples serve as the targets for our attribution analysis\footnote{In addition to attribution on the generated set, we include results for attribution on the test dataset in Appendix~\ref{app:additional-results}.}.

\subsection{Validity: Approximating Causal Attribution with TDA}\label{ssec:causal-attribution-performance}

Our first goal is to evaluate if the practical implementations of Generative Content ID via TDA methods faithfully approximates the causal attribution that requires expensive model retraining. 

\vpara{Evaluation Metric.} The ``gold standard'' for causal attribution is the leave-one-out influence score defined in Eq.~(\ref{eq:influence}). Ideally, we would retrain the model $n$ times (removing one training point each time) to calculate exact scores, and then compute the Spearman's rank correlation between our TDA approximations $\{\hat{I}(m_i, \hat{m})\}_{i=1}^n$ and the ground truth scores $\{I(m_i, \hat{m})\}_{i=1}^n$~\citep{koh2017understanding}.

However, obtaining all the leave-one-out scores via repeated model retraining is computationally infeasible for large datasets. Following~\citet{ilyas2022datamodels} and \citet{park2023trak}, we adopt an approximated version of this rank correlation metric. Instead of removing single points, we randomly generate a collection of subsets $U \subseteq 2^S$, where $2^S$ is the power set of $S$. For each subset $S' \in U$, we retrain a model on $S\setminus S'$ to obtain the ground truth influence of that subset, i.e., slightly abusing the notation, $I(S', \hat{m}) := f(\hat{m}, h_S) - f(\hat{m}, h_{S\setminus S'})$. We compare this against the sum of TDA scores for that subset: $\hat{I}(S', \hat{m}) := \sum_{m\in S'} \hat{I}(m, \hat{m})$. A high rank correlation between these subset-level scores indicates the TDA method faithfully captures the model's retraining behavior. The rank correlation ranges from -1 to 1, where 1 indicates perfect alignment.

\vpara{Results.} Table~\ref{table:spearman-rank-corr} reports the average Spearman's rank correlations across all generated segments. We observe that Generative Content ID, implemented via TDA methods (TRAK and LoGra), achieves significant positive correlations ($p$ < 0.001) with the ground-truth retraining scores, vastly outperforming the random baseline. Notably, LoGra achieves superior performance (0.410 on \aria\ ). These results confirm that TDA is a viable, scalable proxy for causal attribution. Consequently, we utilize LoGra as the implementation for Generative Content ID in the remainder of this paper.

\begin{table}[h]
\centering
\caption{The average Spearman's rank correlation between different TDA methods and retraining. ``Random'' refers to a baseline that employs random attribution scores. ``TRAK''~\citep{park2023trak} and ``LoGra''~\citep{choe2024your} refer to the two TDA methods. (*** p < 0.001, ** p < 0.01, * p < 0.05)} \label{table:spearman-rank-corr}
\resizebox{0.8\textwidth}{!}{%
\begin{tabular}{@{}l|l|l|l|l@{}}
\toprule
                                      & \#Train, \#Generation & Random & TRAK  & LoGra \\ \midrule
MAESTRO + Music Transformer           & 5000, 178             & 0.009  & $0.301^{***}$ & $0.354^{***}$ \\ \midrule
\aria\ \ + Anticipatory Music Transformer & 28000, 500            & 0.004  & $0.366^{***}$ & $0.410^{***}$ \\ \bottomrule
\end{tabular}%
}
\end{table}

\subsection{Comparison with Legal Proxies: Causal Attribution vs. Perceived Attribution}\label{ssec:perceived-causal-experiment}

Current copyright practice often relies on \emph{perceived similarity} as a proxy for infringement. Here, we investigate whether this legal proxy effectively captures the technical reality of how AI training mechanics utilize training data, by comparing \emph{causal attribution} against \emph{perceived attribution}.

Ideally, we would directly compare the ``gold standards'' of both approaches: leave-one-out retraining for causal attribution and human evaluations for perceived attribution. However, these two approaches are computationally or economically expensive. For retraining, we can only afford subset-level retraining; for human evaluation, we can only afford human annotations of a few hundred pairs of samples. Therefore, we leverage efficient approximations, TDA methods and computational similarity metrics, respectively for causal attribution and perceived attribution, to perform the comparison. 

\subsubsection{Comparison 1: Computational Similarity vs. Retraining}
We compare computational similarity against the ``gold standard'' subset retraining in terms of the Spearman's rank correlation evaluation metric defined in Section~\ref{ssec:causal-attribution-performance}, where we replace the TDA scores with the similarity scores. We evaluate three computational similarity metrics, MERT, CLAP and PMI. 

\vpara{Result 1: Computational similarity fails to predict causal attribution.} In Table~\ref{table:comp-sim-lds}, the results show that all three computational similarity metrics have correlations close to zero. As a reference, the TDA methods, TRAK and LoGra, achieved correlations of 0.366 and 0.410 respectively on the same setting. This suggests that, on average, the computational similarity metrics fail to capture the causal influence of the training samples.

\begin{table}[h]
\centering
\caption{The average Spearman's rank correlation between computational similarity metrics and retraining. (*** p < 0.001, ** p < 0.01, * p < 0.05)}\label{table:comp-sim-lds}
\resizebox{0.7\textwidth}{!}{%
\begin{tabular}{@{}l|l|l|l|l@{}}
\toprule
                                      & \#Train, \#Generation & MERT  & CLAP  & PMI   \\ \midrule
\aria\ \ + Anticipatory Music Transformer & 28000, 500            & 0.016 & $0.014$ & $0.068^{**}$ \\ \bottomrule
\end{tabular}%
}
\end{table}

\subsubsection{Comparison 2: Computational Similarity vs. TDA}

We further compare computational similarity metrics against a TDA method, LoGra, to obtain a more fine-grained understanding of the relationship between perceived attribution and causal attribution. Here, since both approaches are efficient, we are able to obtain the TDA score and the similarity score for every pair of training sample and generated sample. We rank the training samples by their average TDA scores across generated samples, and then plot the average similarity scores vs. the rank.

\begin{figure}[h]
    \centering
    \includegraphics[width=0.32\linewidth]{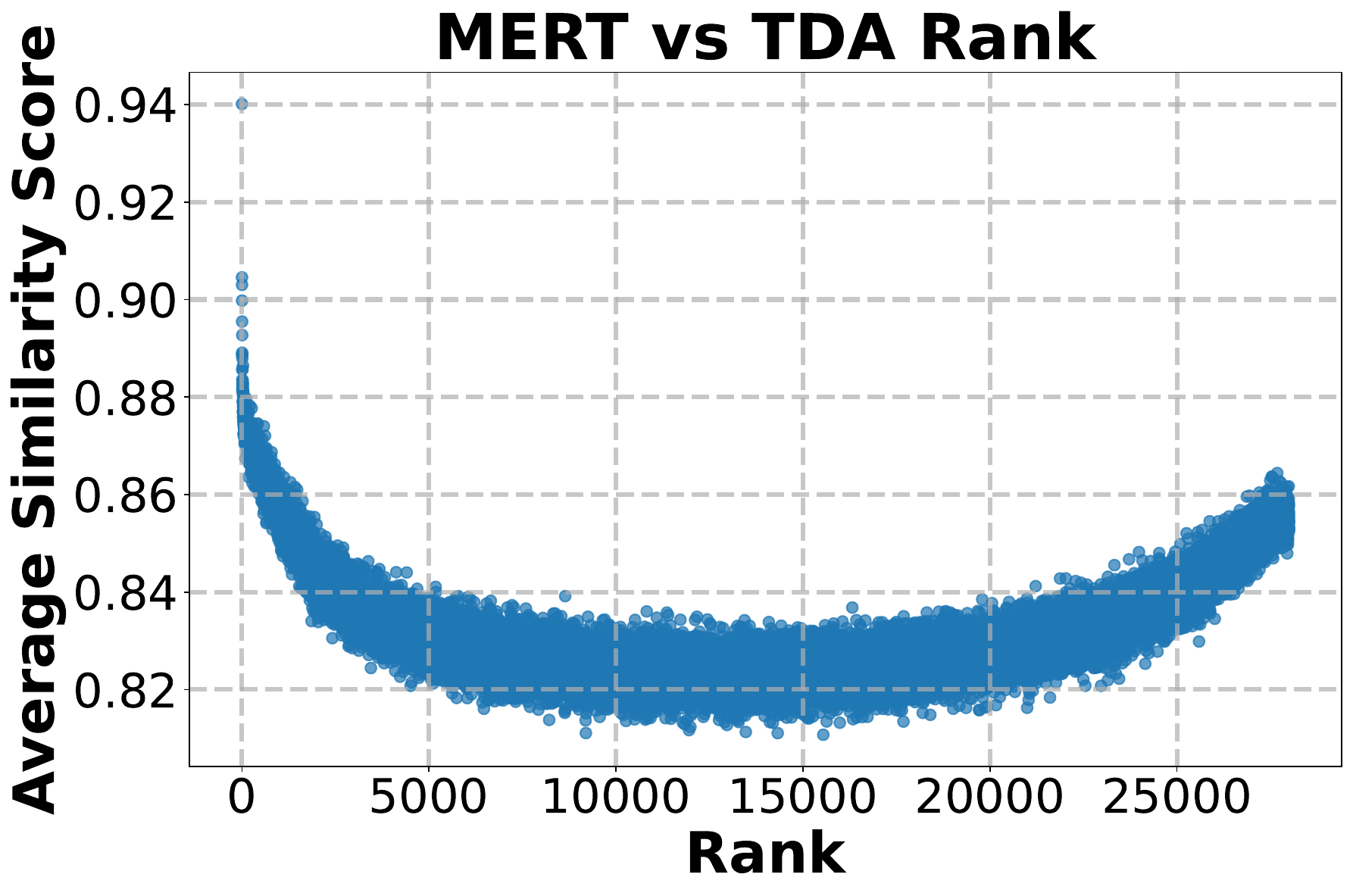}
    \includegraphics[width=0.32\linewidth]{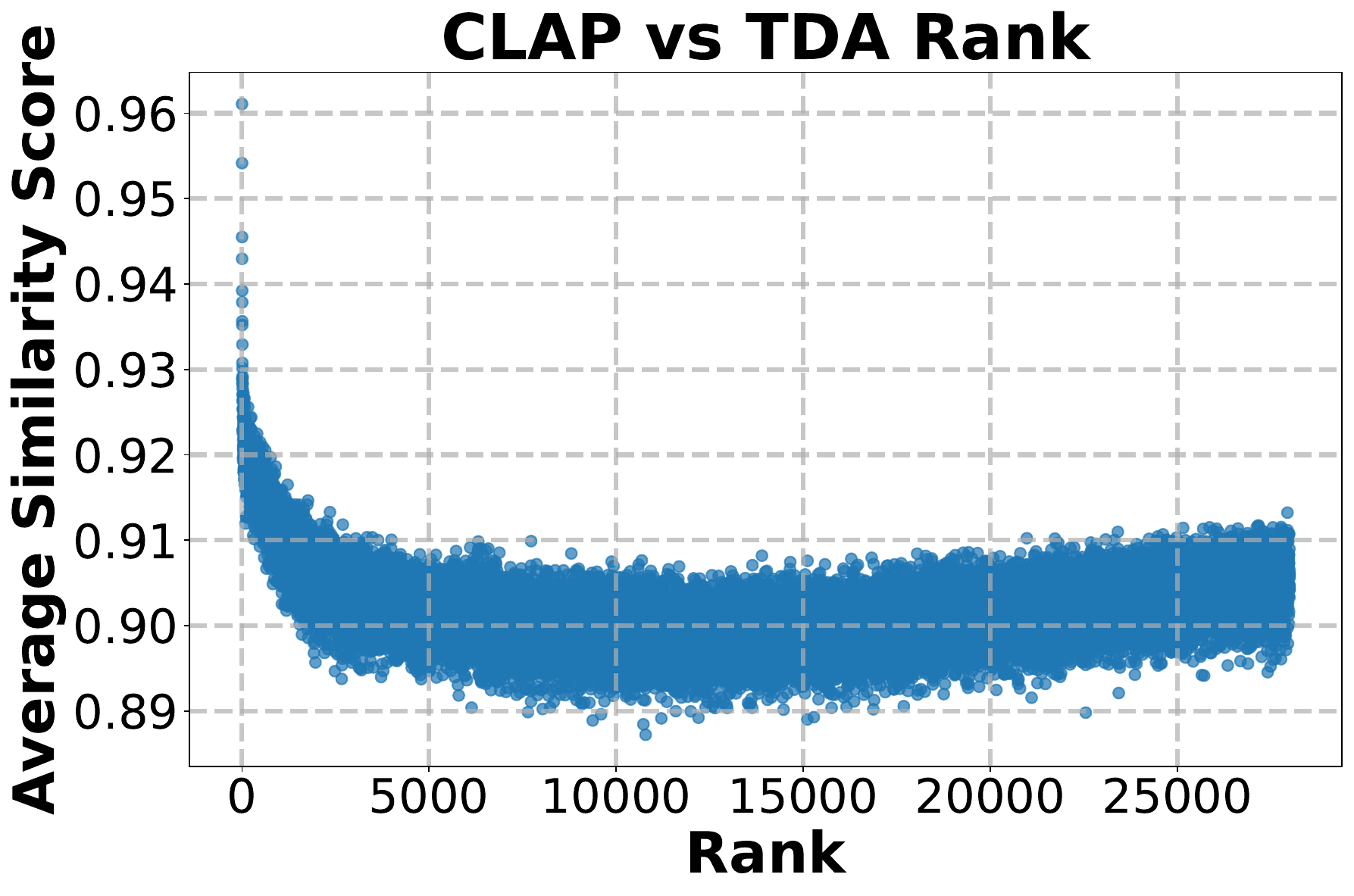}
    \includegraphics[width=0.32\linewidth]{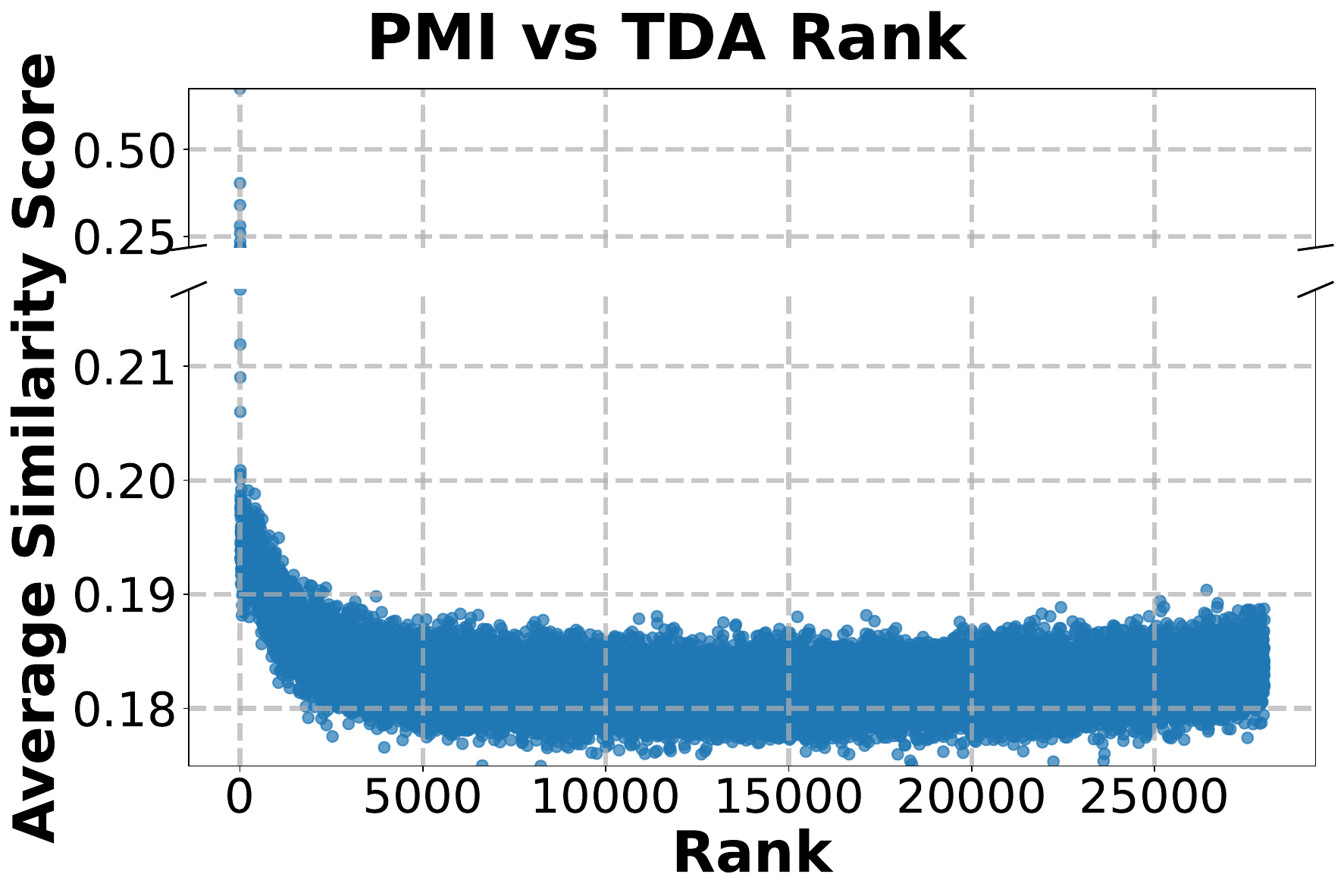}
    \caption{Average similarity score vs. rank by average TDA scores: MERT (left), CLAP (middle), and PMI (right). The x-axis represents the rank of each training sample based on the average TDA scores across generated samples, and the y-axis represents the average similarity scores between each training sample and all the generated samples.}
    \label{fig:tda-comp-similarity}
\end{figure}

\vpara{Result 2: Computational similarity aligns poorly with TDA on average, but they agree on the most influential training samples.} Figure~\ref{fig:tda-comp-similarity} presents the relationship between computational similarity metrics and the TDA (LoGra) rank, where each point in the plot represents a training music sample. This fine-grained plot reveals that while computational similarity aligns poorly with TDA on average, interestingly, they agree on the most influential training samples. Specifically, for all three computational similarity metrics, the top few training samples ranked by TDA (the left-most samples on the plot) have significantly higher average similarity scores than the remaining samples. 

Another interesting finding is that the plots have a U-shape\footnote{The U-shape is most salient for MERT while being less obvious for CLAP and PMI. But the average similarity scores of the bottom-ranked samples are also slightly higher than average for CLAP and PMI.}, which suggests that some training samples that are perceived as similar to the generated sample may actually have high \emph{negative} influence on its generation.

\subsubsection{Comparison 3: TDA vs. Human Evaluation}\label{sssec:tda-vs-human-evaluation}

We conducted an IRB-approved human listening study to compare human evaluation of perceived similarity with the TDA method. In the study, participants are asked to listen to pairs of music (one generated, one training) and rate their similarity on a 1-5 scale. The music samples presented in the study come from the following procedure: (1) we first randomly sample 20 pieces of generated music; and then for each generated sample, (2) we select five training samples that are respectively ranked $1$st, $10$th, $100$th, $n/2$th, $(n-100)$th, and $n$th in terms of the TDA scores of training samples on this specific generated sample; (3) we select another five training samples at the same ranks but measured by computational similarity scores (MERT). The inclusion training samples ranked by computational similarity scores serves as a sanity check between the computational similarity metric and human evaluation. More detailed information about the human study can be found in Appendix~\ref{app:human-studies}.

\begin{figure}[h]
    \centering
    \begin{subfigure}[b]{0.48\linewidth}
        \centering
        \includegraphics[width=\linewidth]{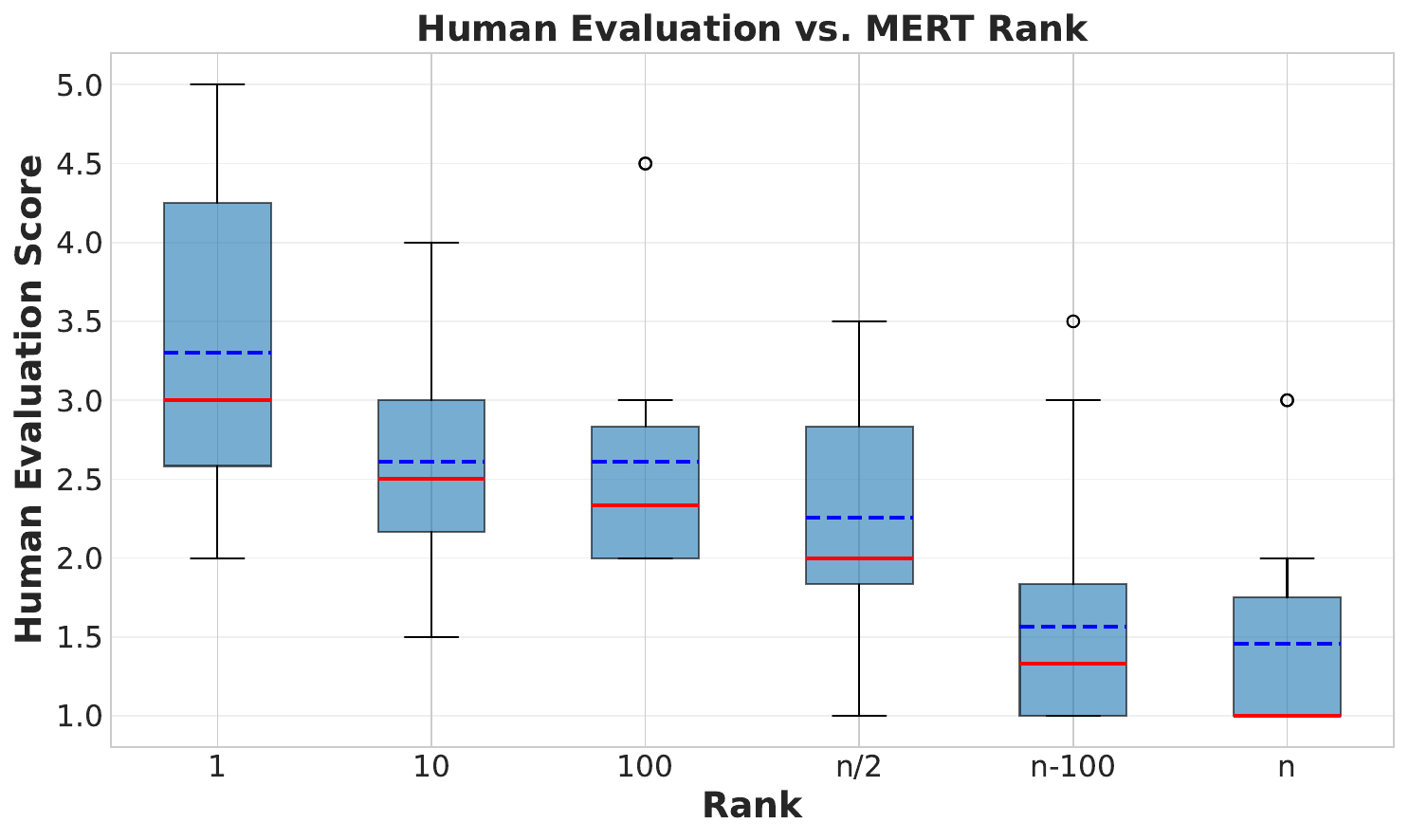}
        \caption{Human ratings vs. MERT rank}
        \label{fig:similarity-human}
    \end{subfigure}
    \hfill
    \begin{subfigure}[b]{0.48\linewidth}
        \centering
        \includegraphics[width=\linewidth]{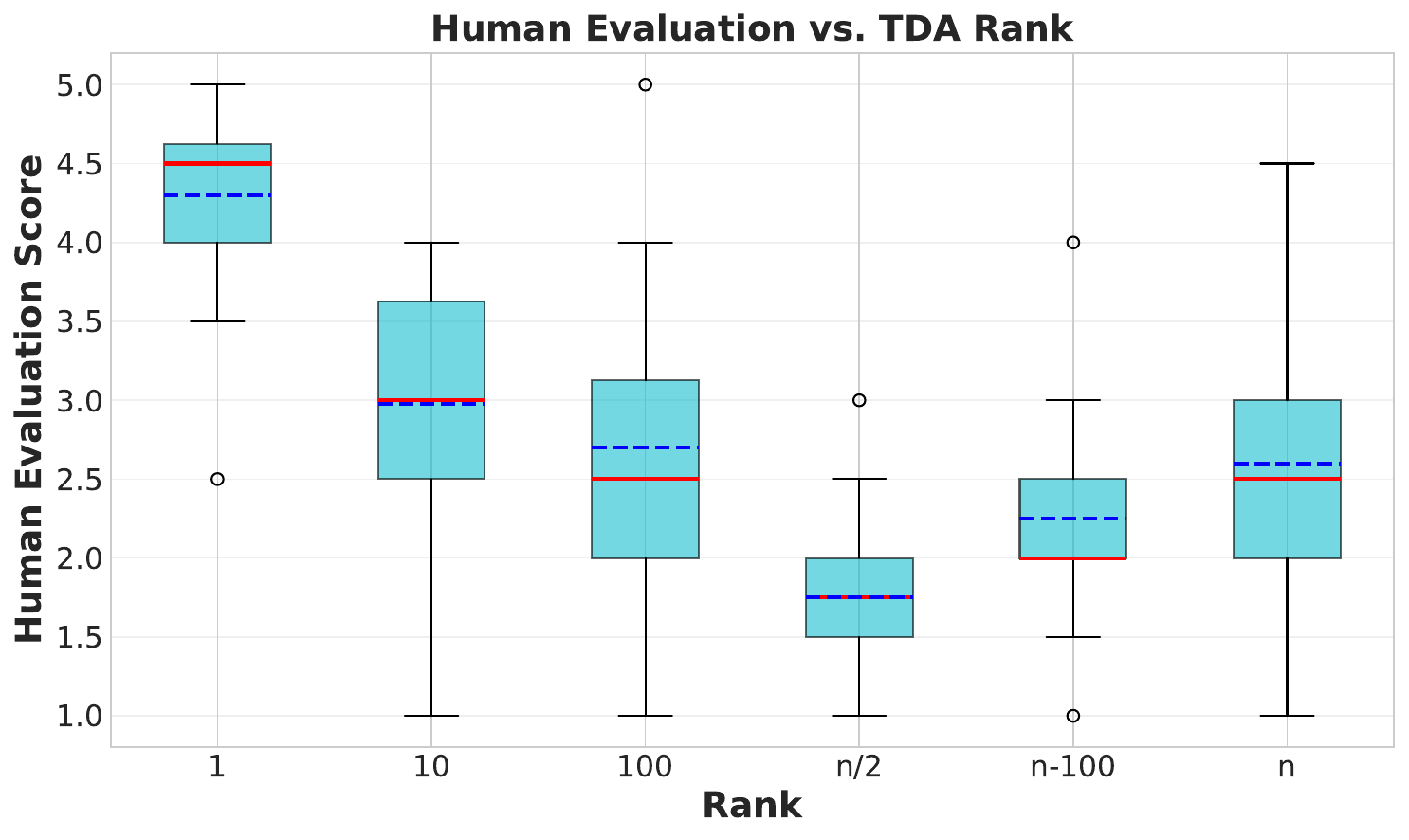}
        \caption{Human ratings vs. LoGra rank}
        \label{fig:tda-human}
    \end{subfigure}
    \caption{Box plots comparing human evaluation with computational similarity or TDA. The x-axis represents groups of training samples divided by increasing rank according to computational similarity scores (MERT) or TDA scores (LoGra) calculated on the generated samples. The y-axis represents the similarity scores between the pair of training and generated samples rated by human participants. The red line indicates the median, and the blue dotted line indicates the mean.}
    \label{fig:tda-similarity-human}
\end{figure}

\vpara{Result 3: Human evaluation shows a similar trend as computational similarity.} Figure~\ref{fig:tda-similarity-human} shows the results of the human listening study. Overall, the human ratings of similarity demonstrate a similar trend as the similarity metric, MERT. Figure~\ref{fig:similarity-human} confirms that the human ratings correlate well with the MERT scores. Figure~\ref{fig:tda-human} further demonstrates a similar U-shaped curve as observed in Figure~\ref{fig:tda-comp-similarity}. 
This result reinforces the validity of results 1 and 2 using real human evaluations. 

\subsubsection{The Influence of ``Hidden Influencers''}

Finally, we highlight the potential risks of relying solely on the legal proxies/perceived similarities for royalty distribution. Since TDA and computational similarity are not linearly correlated, there exists a class of ``Hidden Influencers'': music samples that exhibit \emph{low} perceived similarity yet possess \emph{high} causal attribution scores.

\begin{figure}[h]
    \centering
    \includegraphics[width=0.6\linewidth]{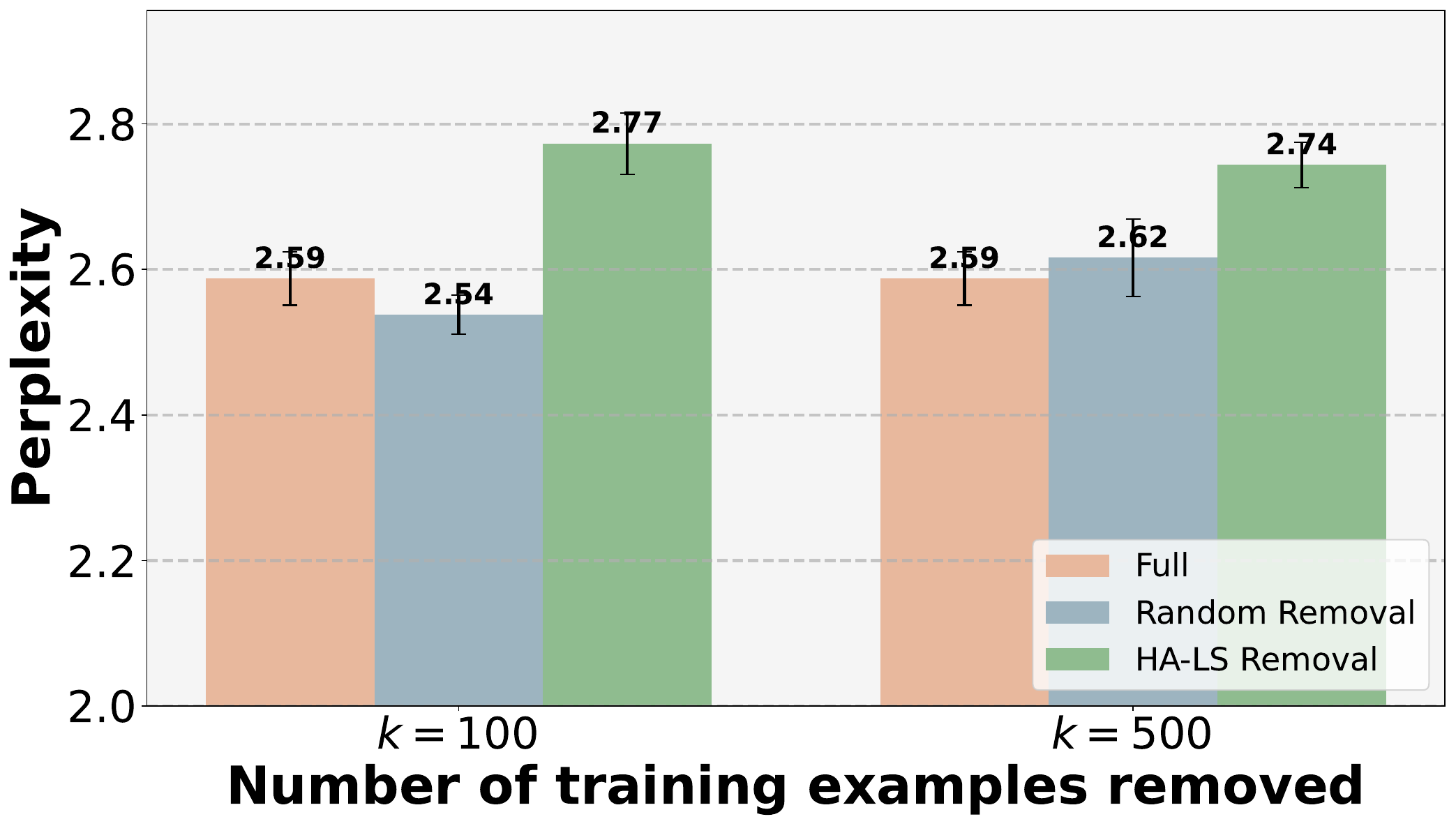}
    \caption{Perplexity of the model on a generated sample when removing $k$ training samples. Full: no removal; Random Removal: removing randomly selected samples; HA-LS Removal: removing HA-LS samples. The error bar shows the standard error of the mean over 5 independent runs.}
    \label{fig:data-removal}
\end{figure}

In Figure~\ref{fig:data-removal}, we simulate the removal of these High-Attribution-Low-Similarity (HA-LS) samples\footnote{Please refer to Appendix~\ref{sec:causal-similarity-removal} for details on the selection of the HA-LS samples.}. Retraining results demonstrate that removing these ``dissimilar'' samples significantly increases the perplexity (i.e., decreases the likelihood) of the generated music. This empirically demonstrates that perceived similarity is an insufficient metric for attribution: if the system relied only on similarity, the creators of these critical HA-LS samples would go uncompensated despite their substantial contribution to the model's generation.

\vpara{Summary of Section~\ref{ssec:perceived-causal-experiment}:} Our investigation reveals a complex, non-linear relationship between legal proxies/perceived attribution and causal attribution. While computational similarity and human evaluation align with causal attribution for the most influential training samples, they fail to capture the broader spectrum of data contributions. Crucially, our experiments on HA-LS samples demonstrate that a royalty framework relying solely on perceived attribution would systematically under-compensate a class of ``hidden influencers.'' Such a misalignment risks undermining the incentive for these rights holders to contribute their works to training datasets, ultimately degrading the long-term quality of generative AI models.

\subsection{Economic Impact Simulation}\label{ssec:economic-impact}

While Sections 4.2 and 4.3 establish the technical validity of Generative Content ID, raw attribution scores alone do not dictate economic outcomes. The translation of these technical scores into actual financial payouts requires a specific royalty distribution mechanism. This mechanism acts as a ``tunable knob'' for platform managers, allowing them to shape the economy of the platform. While \citet{zhang2025fairshare} recently explores how to use influence function for fair pricing practice in LLM data market using a game-theoretical model, the pricing mechanism for loyalty distribution remains unexplored. In this section, we address this gap by providing a preliminary economic simulation of differing baseline pricing strategies. 

More specifically, we focus on income inequality in the simulation. This focus is critical because the digital music economy is historically prone to ``winner-takes-all'' dynamics, where a fraction of ``superstars'' capture the vast majority of revenue \citep{alaei2022revenue}. In an AI-driven ecosystem, this risk is amplified: if the ``long tail'' of niche creators who provide the essential diversity for robust model training, are not equitably compensated, they may exit the market. This would lead to data homogenization and cultural erosion, ultimately degrading the generative models themselves. Therefore, we are interested in seeing whether certain pricing mechanisms may help reduce the inequality gap. 

Given the attribution scores, we simulate royalty distribution mechanisms to understand their impact on income inequality. We consider three distribution mechanisms:
\begin{itemize}
\item \emph{Top-$K$ Uniform}: Royalties are split equally among the top $K$ influential training points.
\item \emph{Top-$K$ Proportional}: Royalties are split proportionally to the attribution scores among the top $K$.
\item \emph{Top-$K$ Rank-Weighted}: Royalties are split according to weights assigned as $1/r$ (where $r$ is the rank).
\end{itemize}

\vpara{Income Inequality Under Different Royalty Distribution Mechanisms.}
We conduct a simulation using the \aria\ \  + Anticipatory Music Transformer setting. We aggregate royalties across all 500 generated samples to the 28,000 training points. Table~\ref{table:royalty-allocation} reports income inequality metrics for different distribution mechanisms and varying values of $K$. 

\begin{table}[h]
\centering
\caption{Inequality metrics for different royalty distribution mechanisms. Gini coefficient ranges from 0 (perfect equality) to 1 (maximum inequality). ``Top $X$\% Share'' indicates the cumulative royalty share received by the top $X$\% of copyright holders.}
\label{table:royalty-allocation}
\resizebox{\textwidth}{!}{%
\begin{tabular}{@{}l|c|c|c|c|c|c|c|c|c|c|c|c@{}}
\toprule
& \multicolumn{4}{c|}{Gini} & \multicolumn{4}{c|}{Top 1\% Share} & \multicolumn{4}{c}{Top 10\% Share} \\ \cmidrule(lr){2-5} \cmidrule(lr){6-9} \cmidrule(lr){10-13}
Method & $K=1$ & $K=10$ & $K=100$ & $K=500$ & $K=1$ & $K=10$ & $K=100$ & $K=500$ & $K=1$ & $K=10$ & $K=100$ & $K=500$ \\ \midrule
Top-$K$ Uniform & 0.983 & 0.864 & 0.466 & 0.262 & 0.596 & 0.130 & 0.043 & 0.026 & 1.000 & 0.695 & 0.282 & 0.196 \\ \midrule
Top-$K$ Proportional & 0.983 & 0.876 & 0.513 & 0.311 & 0.596 & 0.147 & 0.051 & 0.032 & 1.000 & 0.749 & 0.317 & 0.222 \\ \midrule
Top-$K$ Rank-Weighted & 0.983 & 0.914 & 0.704 & 0.563 & 0.596 & 0.227 & 0.137 & 0.108 & 1.000 & 0.876 & 0.560 & 0.457 \\ \bottomrule
\end{tabular}%
}
\end{table}

Several patterns emerge from the results. First, all methods exhibit high inequality when $K=1$, as only a single training music training sample is rewarded for each generated music sample. As $K$ increases, inequality decreases across all mechanisms. Second, the choice of royalty distribution method significantly affects income inequality: ``Top-$K$ Uniform'' produces the most equal outcomes, while ``Top-$K$ Rank-Weighted'' concentrates royalties among the highest-ranked contributors. At $K=500$, the top 1\% of training points receive 2.6\% of royalties under ``Top-$K$ Uniform'' but 10.8\% under ``Top-$K$ Rank-Weighted''. These findings suggest that different royalty distribution mechanisms and parameters such as $K$ can serve as tunable knobs to balance between rewarding the most influential contributors and ensuring broader participation in compensation. 

Ultimately, this analysis underscores that achieving a sustainable AI music economy requires more than just accurate technical attribution; it demands deliberate economic design. The simulation results demonstrate that the specific choice of royalty distribution mechanism serves as a powerful governance tool, determining whether the platform prioritizes a concentrated ``winner-takes-all'' meritocracy or a broader, more inclusive creator ecosystem. 
However, the optimal royalty distribution mechanism may depend on the context of the applications, which is left for future exploration.

\section{Discussion and Conclusion}\label{sec:conclusion}

\subsection{Managerial Implications}\label{sec:discussion}

In this section, we discuss the implications of the novel royalty framework proposed in this paper for policymakers, creators, and platforms in the AI music ecosystem.

\vpara{Policymakers and Regulators.} Our research provides a ``governance-by-design'' roadmap for AI music regulation. Current legislative efforts, such as Tennessee's ELVIS Act~\citep{elvis_act_2024} or the proposed federal Copyright Disclosure Act~\citep{ai_copyright_disclosure_act_2024}, focus largely on preventing unauthorized use or mandating static disclosure. While necessary, these measures do not solve the economic problem of valuation. Rather than banning technologies or relying solely on transparency, regulators can mandate the implementation of faithful royalty frameworks that ensure sustainable compensation. Crucially, our empirical analysis reveals that \emph{perceived similarity} is an insufficient proxy for attribution (Section~\ref{ssec:perceived-causal-experiment}). This has a profound regulatory implication: effective enforcement cannot rely on external ``black-box'' audits of generated output (Does it sound like Artist X?). 
Instead, because faithful attribution requires access to the model's training dynamics, regulation plays an essential role in mandating the ``white-box'' transparency necessary to implement these algorithms. Failing to compensate ``Hidden Influencers'' risks eroding the incentive for foundational data creation, ultimately harming the collective creativity.

\vpara{Artists and Record Labels.} Major record labels have explicitly called for attribution systems analogous to YouTube's Content ID to identify when copyrighted material is incorporated into AI outputs~\citep{wsj2025_licensing}. As demonstrated in Section~\ref{ssec:causal-attribution-performance}, Generative Content ID fulfills this market need, offering a technically faithful mechanism for tracking data usage. Adopting this framework fundamentally shifts the economic model for rights holders. Currently, the primary model is one-time licensing (selling data for a flat fee). A royalty-based framework transforms training data from a fixed asset into a recurring revenue stream. This allows artists to retain an equity-like stake in the AI economy: as the platform's revenue grows, the creator's compensation grows proportionally, ensuring long-term alignment between human creativity and machine innovation.

\vpara{AI Platform Managers.} For AI platform managers, the specific design of the royalty distribution mechanism acts as a critical strategic lever. As shown in Section~\ref{ssec:economic-impact}, adjusting the distribution parameters (e.g., $K$) allows platforms to tune the level of income distribution among their data contributors. Exploration of optimal royalty distribution requires better modeling of key stakeholders' empirical behaviors, which remains an open problem for future work.

\subsection{Conclusion}

In conclusion, this paper has explored the intricate landscape of copyright challenges posed by generative AI, with a particular emphasis on the music industry. We highlight that managing the valuation of AI-generated content is not merely a legal challenge, but a technical one that requires robust infrastructure to ensure sustainable innovation.

Our work bridges this gap by synthesizing the economic logic of existing digital music platforms with the technical capabilities of training data attribution. We propose a royalty framework that operationalizes the measurement of indirect consumption, addressing the core challenge of tracing generated outputs back to training data. This study serves as a foundational prototype for a computational copyright solution, offering a viable blueprint for a fair and scalable economy in the generative AI era.


\newpage
\appendix

\section{A Primer on the Concepts of Music Royalties}\label{concept-of-music-royalties}
It is essential to familiarize ourselves with the fundamental concepts and a few major types of music royalties that are prevalent in the industry.

\vpara{Streaming Royalty.} Streaming royalty arises when music is played on platforms like Spotify, Apple Music, or Amazon Music. Here, copyright owners  earn from the platform's revenue, which could be from subscriptions or advertisements. This concept extends to video platforms like YouTube. When copyrighted music is used in the videos, the copyright owners can monetize those videos through advertisements. Streaming royalty generates recurring revenue shares from the platforms.

\vpara{Synchronization License.} Synchronization license usually applies to the use of music in visual media like movies, TV shows, advertisements, video games, and other types of videos (including YouTube videos). The copyright owner grants a license for the music to be synchronized with specific visual content, usually for a one-time payment. 

\vpara{Mechanical Royalty.} Mechanical royalties are earned when a song is reproduced\footnote{This category also includes the case when a song is reproduced in physical forms, such as piano rolls (where the term ``mechanical'' originates), phonograph records, CDs or vinyl records. But these forms are irrelevant to the purpose of this paper.}, such as when a song is downloaded. Mechanical royalties also apply to derivatives of the music. For example, when a DJ or producer creates a remix, they need to obtain mechanical licenses for the songs they are using. Every time the remix is sold, streamed, or otherwise distributed, mechanical royalties are due to the original copyright owners of the songs used in the remix.

It is worth noting that these royalty categories, although insightful for understanding historical practices and industry folklore, do not constitute formal, legal, or mutually exclusive classifications. Furthermore, the outlined descriptions of music royalties are presented at a high level and are considerably simplified. The implementation of royalty models in a real-world context is complex and varies significantly across different platforms. The process of calculating and distributing these royalties involves multiple stakeholders, such as music creators, record labels, and streaming platforms, each with their own agreements and interests. This complexity extends to how revenue is generated—--whether through advertisements, subscriptions, or a combination of both.

\section{Detailed Stakeholder Description}\label{appendix:detailed-stakeholder-description}
\begin{enumerate}[itemsep=0mm, topsep=-2pt, partopsep=-2pt, parsep=0pt]
    \item \textbf{Artists and Creators:} Musicians, songwriters, and producers who create the content streamed on Spotify.
    \item \textbf{Record Labels and Music Publishers:} Organizations that own the copyrights to music recordings and compositions.
    \item \textbf{Music Rights Societies and Collecting Agencies:} Organizations responsible for collecting royalties and distributing them to copyrights owners.
    \item \textbf{Listeners and Subscribers:} The end-users whose subscription fees and advertising views generate revenue.
    \item \textbf{Advertisers:} Companies that pay Spotify to advertise on its free-tier platform.    
\end{enumerate}

\section{More Details of Experimental Setup}\label{more-details-of-experimental-setup}

\vpara{Anticipatory Music Transformer trained on \aria\ .}
For this experiment, we fine-tune an Anticipatory Music Transformer (AMT) (small) on a commercial symbolic-music dataset, \aria\ . The model is a GPT-2-style decoder-only Transformer with 12 layers, 12 attention heads, hidden size 768, context length 1024, GELU activations, dropout rate 0.1 on attention, embeddings, and residual connections, and a total vocabulary of 55{,}028 tokens. We adopt the arrival-time encoding used in AMT, which represents each musical event as a context-free triplet $(t_i, d_i, n_i)$ of discrete tokens. The arrival-time token $t_i$ denotes the absolute onset time of the event within the current context window and is obtained by quantizing continuous time in 10\,ms steps into 10{,}000 discrete values covering the range $[0,100)$ seconds. The duration token $d_i$ encodes the note length in seconds, quantized in 10\,ms steps into 1{,}000 possible duration values up to 10 seconds. The note token $n_i$ indexes an instrument-pitch pair: MIDI pitches $p \in \{0,\dots,127\}$ follow the standard 12-tone scale, instrument classes $k \in \{0,\dots,128\}$ follow the MIDI program numbers, and we map each pair to a single integer $n_i = 128k + p$, yielding 16{,}512 distinct instrument--pitch categories. Concatenating these triplets produces a sequence $x_{1:3N}$ with $x_{3i-2} = t_i$, $x_{3i-1} = d_i$, and $x_{3i} = n_i$. For all \aria\ \ experiments, we use sequences of length 1024 tokens, where the first token is a special start-of-sequence marker for autoregressive conditioning followed by 1023 arrival-time tokens.

After preprocessing \aria\ \  with this tokenizer, we obtain 194{,}343 fixed-length samples. From these, we construct a subset of 28{,}000 sequences that are compatible with our anticipatory setup and use this subset as the training set. We randomly select 500 sequences from the remaining data to serve as our test set. We initialize from the official 124M-parameter pretrained AMT checkpoint and fine-tune it on this \aria\ \  subset with the standard next-token cross-entropy objective for 5 epochs, using a batch size of 16 for both training and evaluation. Optimization is performed with AdamW with learning rate $2\times 10^{-5}$, a linear learning-rate scheduler with 10\% warmup, weight decay $0.01$, and random seed 0; all other hyperparameters remain at their default values in the released configuration. For autoregressive music generation, we use the fine-tuned model to generate continuations conditioned on each of the 500 test sequences as prompts.

\vpara{Music Transformer trained on MAESTRO.}
We use the MIDI and Audio Edited for Synchronous TRacks and Organization (MAESTRO) dataset (v2.0.0)~\citep{hawthorne2018enabling} for our experiments. The dataset contains more than 200 hours of piano performances and is stored in MIDI format. The MIDI format is a widely recognized industry standard for representing musical information. Generally, it encodes music as a sequence of MIDI messages. MIDI messages include attributes like velocity, duration, and pitch key. We refer readers to a tutorial~\citep{de2017understanding} for more details about the MIDI format.

Following the setting of Music Transformer~\citep{huang2018music}, we define a vocabulary set of size 388, which includes ``NOTE ON'' and ``NOTE OFF'' events for 128 different pitches (256 events in total), 100 ``TIME SHIFT'' events for different duration lengths, and 32 ``VELOCITY'' events that set the velocity of the following ``NOTE ON'' events. The raw data is preprocessed as sequences of events from this vocabulary set.

We train a Music Transformer model on the MAESTRO dataset using the official training set in the dataset. During the training, the music sequences are cropped into a fixed length of $p=256$ and the model is trained by maximizing the log-likelihood of these sequences. For generation, we use the official test samples in the dataset as prompt to generate music with length $p=256$. The generated music, denoted as $\left\{\hat{m}_1, \hat{m}_2, \ldots, \hat{m}_{k} \right\}$, will be used to evaluate the data attribution methods. Here $k=178$ is the number of test samples in the dataset.

\section{Details of Human Studies}\label{app:human-studies}
We conducted an IRB-approved human listening study. This section provides detailed information about the study design and participant recruitment.

\vpara{Study Design.} Participants were asked to listen to pairs of music samples and rate their perceived similarity on a 5-point scale. Each participant completed 6 questions, with each question consisting of one generated music sample paired with 6 training samples ranked at different positions according to either TDA or computational music similarity (MERT).

To ensure data quality, one of the 6 questions served as an attention check, where a randomly selected pair consisted of identical music samples. Participants were required to rate this identical pair with a score of 5 to pass the attention check.

In total, each participant rated 36 pairs of music (6 questions × 6 pairs), yielding 30 eligible data pairs per participant (5 questions × 6 pairs) for analysis after excluding the attention check question. The study duration ranged from 30 to 60 minutes per participant.

\vpara{Participants.} We recruited 15 eligible participants, all of whom were college students. All participants successfully passed the attention check question and were included in the final analysis.

\section{Additional Experiment Results on Test Dataset}\label{app:additional-results}
\begin{figure}[h]
    \centering
    \includegraphics[width=0.32\linewidth]{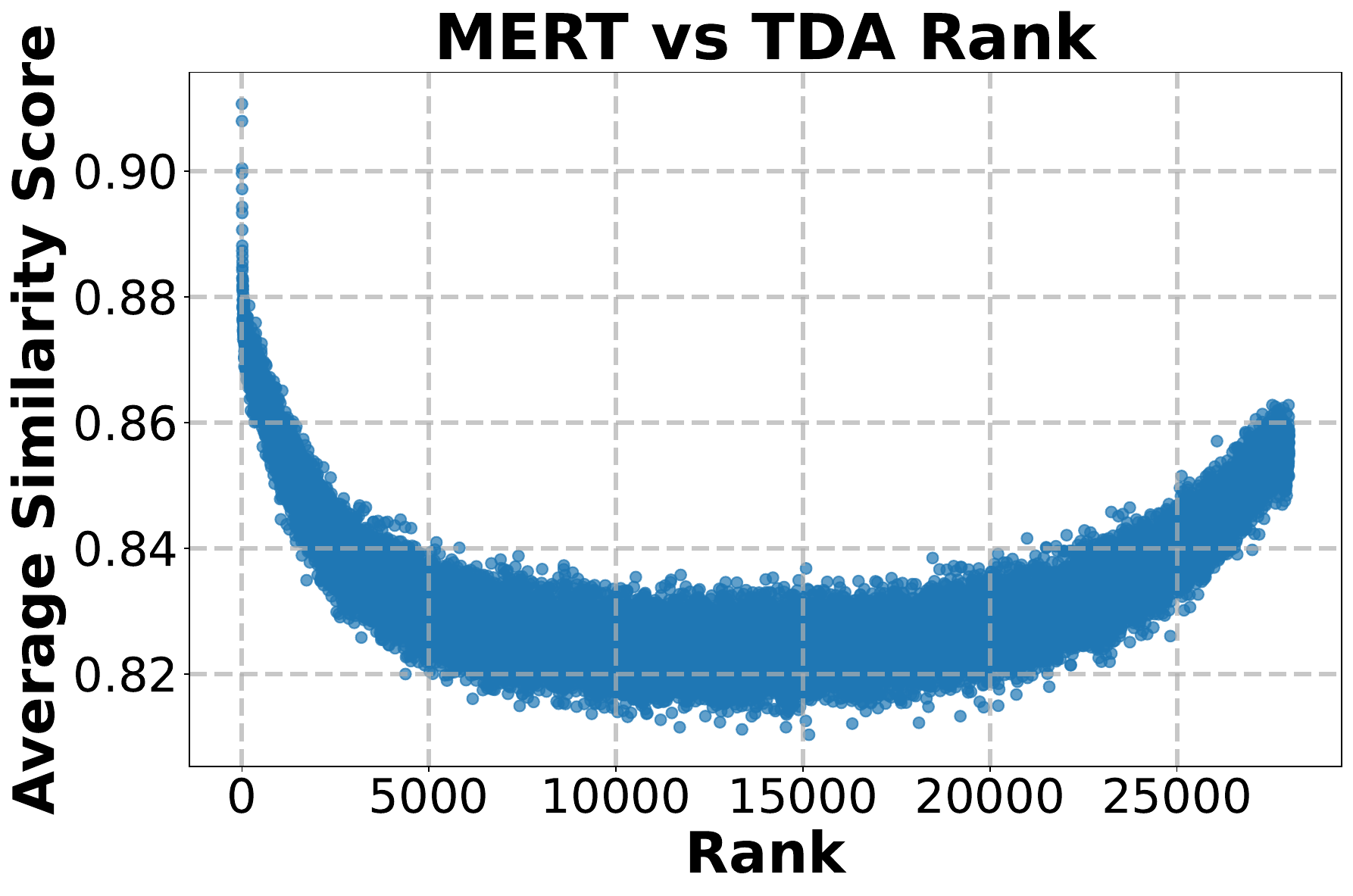}
    \includegraphics[width=0.32\linewidth]{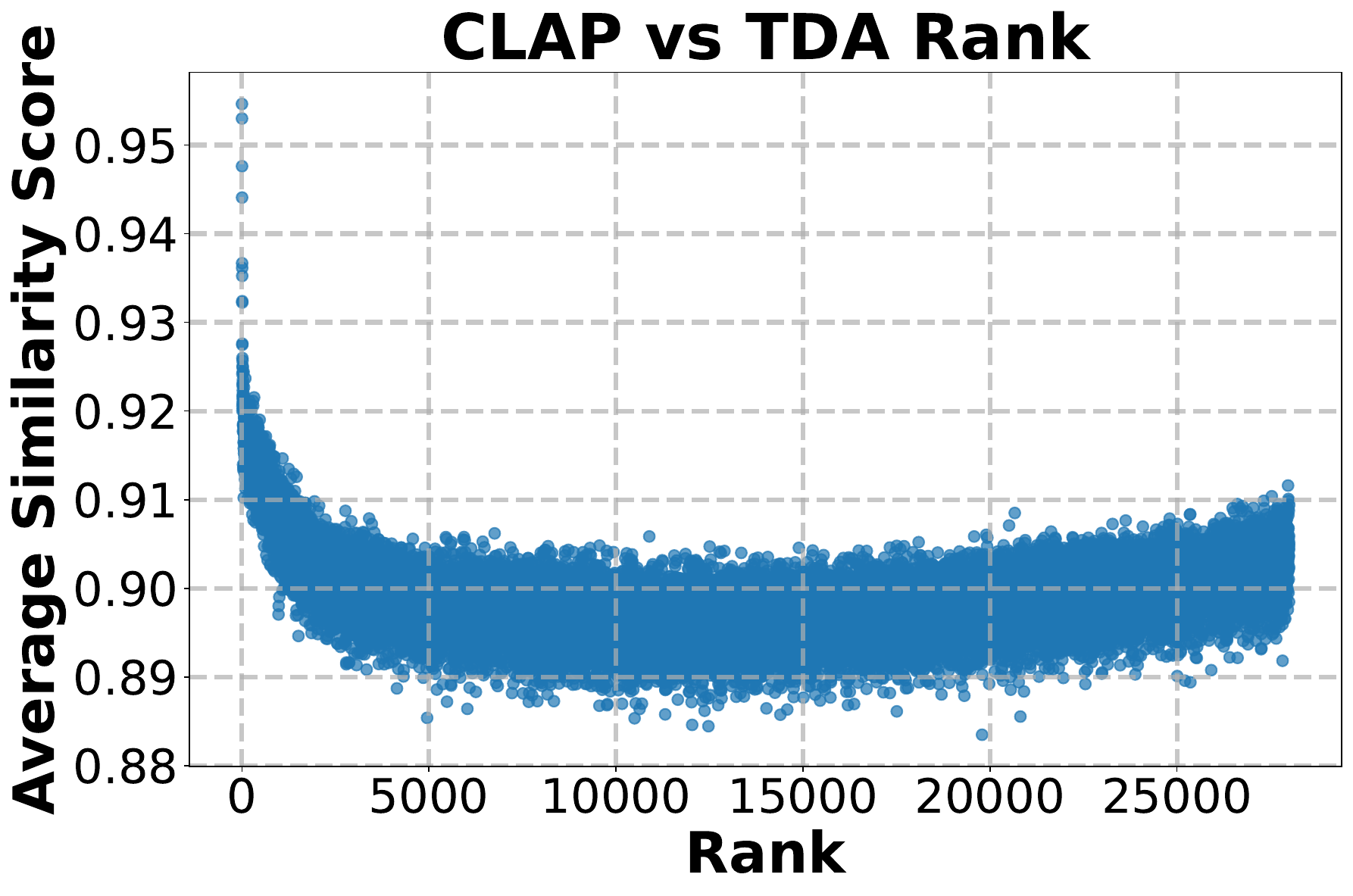}
    \includegraphics[width=0.32\linewidth]{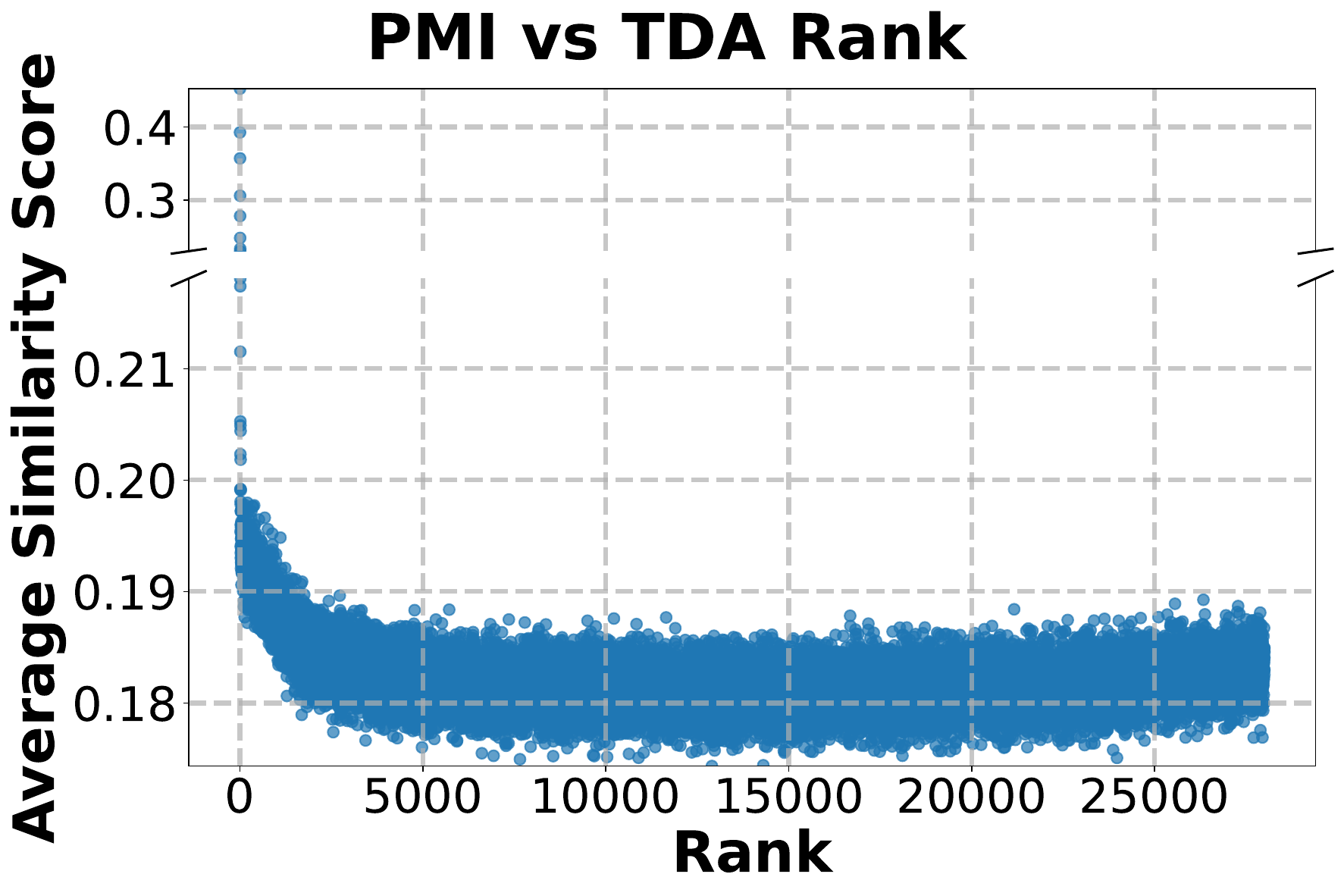}
    \caption{Average similarity score versus TDA rank for MERT (left), CLAP (middle), and PMI (right).}
    \label{fig:similarity-tda-testdata}
\end{figure}

In this section, we present experiments using a test dataset of real music pieces as the target to trace back. Our findings align with those from the experiments using generated music. Table~\ref{table:spearman-rank-corr-testdata} demonstrates that Generative Content ID via gradient-based TDA achieves reasonably high correlation with the ``gold standard'' retraining approach. Table~\ref{table:comp-sim-lds-testdata} shows that computational music similarity methods exhibit poor overall alignment with the causal ``gold standard'' retraining. Figure~\ref{fig:similarity-tda-testdata} illustrates the relationship between TDA and computational music similarity, revealing the same U-shaped pattern observed in the generated music experiments.

\begin{table}[h]
\centering
\caption{The average retraining rank correlation among test music for different data attribution methods. ``Random'' refers to a baseline that employs random attribution scores. (*** p < 0.001, ** p < 0.01, * p < 0.05)} \label{table:spearman-rank-corr-testdata}
\resizebox{0.7\textwidth}{!}{%
\begin{tabular}{@{}l|l|l|l|l@{}}
\toprule
                                      & \#Train, \#Test & Random & TRAK  & LoGra \\ \midrule
\aria\ \  + Anticipatory Music Transformer & (28000, 500)          &     -0.002   & $0.366^{***}$ & $0.418^{***}$ \\ \bottomrule
\end{tabular}%
}
\vspace{5pt}
\end{table}

\begin{table}[h]
\centering
\caption{The average Spearman's rank correlation between computational similarity metrics and retraining. (*** p < 0.001, ** p < 0.01, * p < 0.05)}\label{table:comp-sim-lds-testdata}
\resizebox{0.7\textwidth}{!}{%
\begin{tabular}{@{}l|l|l|l|l@{}}
\toprule
                                      & \#Train, \#Generation & MERT  & CLAP  & PMI   \\ \midrule
\aria\ \  + Anticipatory Music Transformer & 28000, 500            & 0.006 & $0.012^{*}$ & 0.053 \\ \bottomrule
\end{tabular}%
}
\end{table}

\section{Algorithm Specifications}

\subsection{Training Data Attribution Algorithms} \label{sec:training-data-attribution-algorithms}

In this section, we introduce two state-of-the-art TDA methods.

\vpara{Tracing with the Randomly-projected After Kernel (TRAK).} TRAK is proposed by \citet{park2023trak} featured by its effectiveness and scalability. The method introduce random projection to reduce the computational cost. It can be presented as 
\begin{align*}
    \hat{I}_{\rm TRAK}&(m_i, \hat{m}) = \phi_h(\hat{m})(\Phi_h^\top\Phi_h)^{-1}\phi_h^\top(m_i) \cdot q(m_i),
\end{align*}
where $\phi_h(\hat{m})$ and $\phi_h(m_i)$ is the gradient of $f(\hat{m}, h_S)$ and $f(m_i, h_S)$ with respect to the parameter of $h$, $\Phi_h$ is a projected gradient stacked over the training samples $m_i \in S$, $q(m_i)$ is the ``one minus correct-class probability'' of a training data point $m_i$ under model $h$.

For the implementation of attribution score calculation with TRAK~\citep{park2023trak}, we utilize the \texttt{dattri}~\citep{deng2024dattri} toolkit. For the ensemble setting, we use 50 models that are independently trained on 50\% random subset of the training set for Music Transformer experiment and 3 models for Anticipatory Music Transformer.

\vpara{LoGra.} LoGra is proposed by \citet{choe2024your} apply low-Rank gradient projection and factorization to the gradient calculation to further improve the efficiency. The method can be presented as
\begin{align*}
    \hat{I}_{\rm LoGra}&(m_i, \hat{m}) =\\
    &- (Pg_f(\hat{m}))^{\top}(PHP^{\top})^{-1}(Pg(m_i)),
\end{align*}
where the notations are the same as influence function presented in Section~\ref{sssec:scalable-approximation-data-attribution} with an additional projection matrix $P$ on the gradient and hessian matrix. Taking linear layer as an example, LoGra further propose an additional Kronecker-product structure on the projection matrix $P$ as following formula
\begin{align*}
Pg = (P_i \otimes P_o)(x_{i} \otimes g_{x_{o}}) = P_i x_{i} \otimes P_o g_{x_{o}},
\end{align*}
where $g$ is the gradient vector with respect to the parameter, $x_i$ and $x_o$ are the input and pre-activation output of the linear layer, $g_{x_o}$ is the gradient with respect to $x_o$, $P_i$ and $P_o$ are decomposited projection matrix and $\otimes$ is the Kronecker-product. Futhermore, the Hessian matrix can be replaced by empirical Fisher Information Matrix to reduce the computational cost.

For the implementation of attribution score calculation with LoGra~\citep{choe2024your}, we utilize the \texttt{dattri}~\citep{deng2024dattri} toolkit. For the ensemble setting, we use 10 models that are independently trained on full training set for Music Transformer experiment and 1 model for Anticipatory Music Transformer.

\subsection{Computational Music Similarity Methods} \label{sec:music-similarity}
We carry out experiment on three computational music similarity methods.

\vpara{CLAP representation similarity.}
CLAP~\citep{elizalde2023clap} learns joint audio and text representations by contrastive pretraining on audio caption pairs in a shared embedding space.
In our experiments, symbolic music is first rendered to audio and then encoded by the CLAP audio encoder, which we denote as $\phi_{\text{CLAP}}(\cdot)$.
Let $m_i$ be training music segment and $\hat{m}$ be generated music segment.
We compute a cosine similarity matrix $\hat{I}_{\text{CLAP}}$ as
\[
\hat{I}_{\text{CLAP}}(m_i, \hat{m}) = \frac{\phi_{\text{CLAP}}(m_i)^\top \phi_{\text{CLAP}}(\hat{m})}{\lVert \phi_{\text{CLAP}}(m_i)\rVert_2 \lVert \phi_{\text{CLAP}}(\hat{m})\rVert_2}.
\]

\vpara{MERT representation similarity.}
MERT~\citep{li2024mert} is a large scale self supervised music representation model with a 25 layer Transformer encoder for music audio.
We render symbolic music to audio and feed each clip into the pretrained MERT model, then take the hidden representation from the final encoder layer and apply mean pooling over the time dimension to obtain a fixed dimensional embedding, which we denote by $\phi_{\text{MERT}}(\cdot)$.
Let $m_i$ be training music segment and $\hat{m}$ be generated music segment.
We compute a cosine similarity matrix $\hat{I}_{\text{MERT}}$ as
\[
\hat{I}_{\text{MERT}}(m_i, \hat{m}) = \frac{\phi_{\text{MERT}}(m_i)^\top \phi_{\text{MERT}}(\hat{m})}{\lVert \phi_{\text{MERT}}(m_i)\rVert_2 \lVert \phi_{\text{MERT}}(\hat{m})\rVert_2}.
\]

\vpara{Percent melodic identity (PMI).}
PMI~\citep{savage2018quantitative} measures symbolic melodic similarity via global alignment of two pieces.
For each piece we extract a melody line as a sequence of pitch classes, then align two melodies $m_i$ and $\hat{m}$ with a Needleman Wunch algorithm with affine gap penalties: matches of identical pitch classes receive a score of $+1$ mismatches receive $0$ and gaps incur a gap open penalty $c_{\text{open}} = 12$ and a gap extension penalty $c_{\text{ext}} = 6$ following~\citet{savage2018quantitative}.
Let $M(m_i,\hat{m})$ be the number of exact pitch class matches along the optimal alignment and let $|m_i|$ and $|\hat{m}|$ denote the sequence lengths; we define
\[
\hat{I}_{\text{PMI}}(m_i,\hat{m}) = \frac{M(m_i,\hat{m})}{\bigl(|m_i| + |\hat{m}|\bigr) / 2}
\]
which yields a value between $0$ and $1$ that reflects the similarity. In our implementation we compute $\hat{I}_{\text{PMI}}$ for each transposition and taking the maximum value, which better captures similarity between melodies that differ only by a global key shift.

\subsection{High-Attribution-Low-Similarity Samples} \label{sec:causal-similarity-removal}

We introduce how are the HA-LS samples selected in detail and the the perplexity metric in this section. In this experiment, we use LoGra and MERT as the implementation of TDA and computational similarity method.

\vpara{High-Attribution Low-Similarity (HA-LS) Selection.}

Given attribution scores $\hat{I}_{\rm LoGRA}(m_i, \hat{m})$ and computational music similarity scores $\hat{I}_{\rm MERT}(m_i, \hat{m})$ for $m_i \in S$. We first calculate the z-score to get standardized the attribution scores $\hat{I}^*_{\rm LoGRA}(m_i, \hat{m})$ and similarity scores $\hat{I}^*_{\rm MERT}(m_i, \hat{m})$. We then calculate the difference between the standardized attribution scores and similarity scores and select those with largest $\hat{I}^*_{\rm LoGRA}(m_i, \hat{m}) - \hat{I}^*_{\rm MERT}(m_i, \hat{m})$.

\vpara{Perplexity.}
We measure model performance using perplexity on the target generated sample:
\[
\text{PPL} = \exp\left(-\frac{1}{T} \sum_{t=1}^{T} \log P(e_t \mid e_{<t})\right)
\]
where $T$ is the sequence length and $P$ is the model's predicted probability. Higher perplexity indicates that the model assigns lower probability to the generated sequence, suggesting degraded generation capability of that specific sequence.
\end{document}